\documentclass{bmcart}
	
\usepackage{xcolor}
\RequirePackage{natbib}
\usepackage[utf8]{inputenc} %unicode support
\usepackage{graphicx}
\usepackage{multirow}
\usepackage{tabularx}
\colorlet{lightgrey}{lightgray}
\usepackage{url}
\usepackage{amsmath}
\usepackage{multirow}
\usepackage{colortbl}
\usepackage{nccmath}
\usepackage{amsmath,amsfonts,latexsym,amssymb,euscript,xr}
\startlocaldefs
\endlocaldefs

\begin{document}

%%% Start of article front matter
\begin{frontmatter}

\begin{fmbox}
\dochead{Research}


\title{Mining On Alzheimer's Diseases Related Knowledge Graph to Identity Potential AD-related Semantic Triples for Drug Repurposing}

%%%%%%%%%%%%%%%%%%%%%%%%%%%%%%%%%%%%%%%%%%%%%%
%%                                          %%
%% Enter the authors here                   %%
%%                                          %%
%% Specify information, if available,       %%
%% in the form:                             %%
%%   <key>={<id1>,<id2>}                    %%
%%   <key>=                                 %%
%% Comment or delete the keys which are     %%
%% not used. Repeat \author command as much %%
%% as required.                             %%
%%                                          %%
%%%%%%%%%%%%%%%%%%%%%%%%%%%%%%%%%%%%%%%%%%%%%%

\author[
  addressref={aff1},                   % id's of addresses, e.g. {aff1,aff2}
%   corref={},                       % id of corresponding address, if any
%   noteref={},                        % id's of article notes, if any
  email={yi.nian@uth.tmc.edu}   % email address
]{\inits{YN}\fnm{Yi} \snm{Nian}}
\author[
  addressref={aff1},                   % id's of addresses, e.g. {aff1,aff2}
%   corref={},                       % id of corresponding address, if any
%   noteref={},                        % id's of article notes, if any
  email={xinyue.hu.1@uth.tmc.edu}   % email address
]{\inits{XH}\fnm{Xinyue} \snm{Hu}}
\author[
  addressref={aff2},                   % id's of addresses, e.g. {aff1,aff2}
%   corref={aff2},                       % id of corresponding address, if any
%   noteref={},                        % id's of article notes, if any
  email={zhan1386@umn.edu}   % email address
]{\inits{RZ}\fnm{Rui} \snm{Zhang}}
\author[
  addressref={aff1},                   % id's of addresses, e.g. {aff1,aff2}
%   corref={},                       % id of corresponding address, if any
%   noteref={},                        % id's of article notes, if any
  email={Jingna.Feng@uth.tmc.edu}   % email address
]{\inits{JF}\fnm{Jingna} \snm{Feng}}
\author[
  addressref={aff1},                   % id's of addresses, e.g. {aff1,aff2}
%   corref={},                       % id of corresponding address, if any
%   noteref={},                        % id's of article notes, if any
  email={Jingcheng.Du@uth.tmc.edu}   % email address
]{\inits{JD}\fnm{Jingcheng} \snm{Du}}
\author[
  addressref={aff1},                   % id's of addresses, e.g. {aff1,aff2}
%   corref={},                       % id of corresponding address, if any
%   noteref={},                        % id's of article notes, if any
  email={Fang.Li@uth.tmc.edu}   % email address
]{\inits{FL}\fnm{Fang} \snm{Li}}

\author[
  addressref={aff4},                   % id's of addresses, e.g. {aff1,aff2}
%   corref={},                       % id of corresponding address, if any
%   noteref={},                        % id's of article notes, if any
  email={}   % email address
]{\inits{LB}\fnm{Larry} \snm{Bu}}
\author[
  addressref={aff4},                   % id's of addresses, e.g. {aff1,aff2}
%   corref={},                       % id of corresponding address, if any
%   noteref={},                        % id's of article notes, if any
  email={yuzhang@som.umaryland.edu}   % email address
]{\inits{YZ}\fnm{Yuji} \snm{Zhang}}
\author[
  addressref={aff3},                   % id's of addresses, e.g. {aff1,aff2}
%   corref={},                       % id of corresponding address, if any
%   noteref={},                        % id's of article notes, if any
  email={ychen123@upenn.edu}   % email address
]{\inits{YC}\fnm{Yong} \snm{Chen}}
\author[
  addressref={aff1},                   % id's of addresses, e.g. {aff1,aff2}
  corref={aff1},                       % id of corresponding address, if any
  noteref={},                        % id's of article notes, if any
  email={Cui.Tao@uth.tmc.edu}   % email address
]{\inits{CT}\fnm{Cui} \snm{Tao}}

% \author[
%   addressref={aff1,aff2},
%   email={john.RS.Smith@cambridge.co.uk}
% ]{\inits{JRS}\fnm{John RS} \snm{Smith}}

%%%%%%%%%%%%%%%%%%%%%%%%%%%%%%%%%%%%%%%%%%%%%%
%%                                          %%
%% Enter the authors' addresses here        %%
%%                                          %%
%% Repeat \address commands as much as      %%
%% required.                                %%
%%                                          %%
%%%%%%%%%%%%%%%%%%%%%%%%%%%%%%%%%%%%%%%%%%%%%%

\address[id=aff1]{%                           % unique id
  \orgname{School of Biomedical Informatics, University of Texas Health Science Center at Houston}, % university, etc
  \street{7000 Fannin St},                     %
  %\postcode{}                                % post or zip code
  \city{Houston},                              % city
  \cny{TX} }                                 % country

\address[id=aff2]{%
  \orgname{Department of Pharmaceutical Care $\&$ Health System (PCHS) and the Institute for Health Informatics (IHI), University of Minnesota.},
  \street{7-115A Weaver-Densford Hall},
  \postcode{55455}
  \city{Minneapolis},
  \cny{MN}
}

\address[id=aff3]{%
  \orgname{Department of Biostatistics, Epidemiology and Informatics (DBEI), the Perelman School of Medicine, University of Pennsylvania},
  \street{602 Blockley Hall, 423 Guardian Drive},
  \postcode{19104}
  \city{Philadelphia},
  \cny{PA}
}

\address[id=aff4]{%
  \orgname{University of Maryland School of Medicine, Baltimore MD USA},
  \street{655 W Baltimore St S},
  \postcode{21201}
  \city{Baltimore},
  \cny{MD}
}
%%%%%%%%%%%%%%%%%%%%%%%%%%%%%%%%%%%%%%%%%%%%%%
%%                                          %%
%% Enter short notes here                   %%
%%                                          %%
%% Short notes will be after addresses      %%
%% on first page.                           %%
%%                                          %%
%%%%%%%%%%%%%%%%%%%%%%%%%%%%%%%%%%%%%%%%%%%%%%

\begin{artnotes}
%\note{Sample of title note}     % note to the article
% \note[id=n1]{Equal contributor} % note, connected to author
\end{artnotes}

\end{fmbox}% comment this for two column layout

%%%%%%%%%%%%%%%%%%%%%%%%%%%%%%%%%%%%%%%%%%%%%%
%%                                          %%
%% The Abstract begins here                 %%
%%                                          %%
%% Please refer to the Instructions for     %%
%% authors on http://www.biomedcentral.com  %%
%% and include the section headings         %%
%% accordingly for your article type.       %%
%%                                          %%
%%%%%%%%%%%%%%%%%%%%%%%%%%%%%%%%%%%%%%%%%%%%%%

\begin{abstractbox}

\begin{abstract} % abstract
\parttitle{Background} %if any

To date, there are no effective treatments for most neurodegenerative diseases. Knowledge graphs can provide comprehensive and semantic representation for heterogeneous data, and have been successfully leveraged in many biomedical applications including drug repurposing. Our objective is to construct a knowledge graph from literature to study the relations between Alzheimer's Disease (AD) and chemicals, drugs and dietary supplements in order to identify opportunities to prevent or delay neurodegenerative progression.  We collected biomedical annotations and extracted their relations using SemRep via SemMedDB. We used both a BERT-based classifier and rule-based methods during data preprocessing to exclude noise while preserving most AD-related semantic triples. The 1,672,110 filtered triples were used to train with knowledge graph completion algorithms (i.e., TransE, DistMult, and ComplEx) to predict candidates that might be helpful for AD treatment or prevention.  
% train data: 1672110 = 649924 113031 28872
% before BERT 267,376

\parttitle{Results} %if any

Among three knowledge graph completion models, TransE outperformed the other two (MR = 10.53, Hits@1 = 0.28). We leveraged the time-slicing technique to further evaluate the prediction results. We found supporting evidence for most highly ranked candidates predicted by our model which indicates that our approach can inform reliable new knowledge.

\parttitle{Conclusion} %if any

This paper shows that our graph mining model can predict reliable new relationships between AD and other entities (i.e., dietary supplements, chemicals, and drugs). The knowledge graph constructed can facilitate data-driven knowledge discoveries and the generation of novel hypotheses.

\end{abstract}

%%%%%%%%%%%%%%%%%%%%%%%%%%%%%%%%%%%%%%%%%%%%%%
%%                                          %%
%% The keywords begin here                  %%
%%                                          %%
%% Put each keyword in separate \kwd{}.     %%
%%                                          %%
%%%%%%%%%%%%%%%%%%%%%%%%%%%%%%%%%%%%%%%%%%%%%%

\begin{keyword}
\kwd{Alzheimer's Disease}
\kwd{Dietary supplement}
\kwd{Drug Repurposing}
\kwd{Knowledge Graph}
\kwd{Literature Mining}

\end{keyword}

% MSC classifications codes, if any
%\begin{keyword}[class=AMS]
%\kwd[Primary ]{}
%\kwd{}
%\kwd[; secondary ]{}
%\end{keyword}

\end{abstractbox}
%
%\end{fmbox}% uncomment this for twcolumn layout

\end{frontmatter}

%%%%%%%%%%%%%%%%%%%%%%%%%%%%%%%%%%%%%%%%%%%%%%
%%                                          %%
%% The Main Body begins here                %%
%%                                          %%
%% Please refer to the instructions for     %%
%% authors on:                              %%
%% http://www.biomedcentral.com/info/authors%%
%% and include the section headings         %%
%% accordingly for your article type.       %%
%%                                          %%
%% See the Results and Discussion section   %%
%% for details on how to create sub-sections%%
%%                                          %%
%% use \cite{...} to cite references        %%
%%  \cite{koon} and                         %%
%%  \cite{oreg,khar,zvai,xjon,schn,pond}    %%
%%  \nocite{smith,marg,hunn,advi,koha,mouse}%%
%%                                          %%
%%%%%%%%%%%%%%%%%%%%%%%%%%%%%%%%%%%%%%%%%%%%%%

%%%%%%%%%%%%%%%%%%%%%%%%% start of article main body
% <put your article body there>

%%%%%%%%%%%%%%%%
%% Background %%
%%

\section*{Background}

Neurodegenerative diseases are a heterogeneous group of disorders that are characterized by the progressive degeneration of the structure and function of the central nervous system or peripheral nervous system\cite{neurodegenerative}. Common neurodegenerative diseases, such as Alzheimer’s disease(AD) and related dementias (ADRD), are usually incurable and irreversible and difficult to stop. 

AD/ADRD are multi-factorial and complex neurodegenerative diseases characterized by progressive memory loss and severe dementia with neuropsychiatric symptoms \cite{Moya-Alvarado}. An estimated 5.8 million Americans aged 65 and older (12.6$\%$) are living with AD/ADRD in 2020, and this number is projected to reach 13.8 million by 2050 \cite{Duan}. High prevalence of AD/ADRD creates huge medical and social burdens. The total costs for health care, long-term care and hospital services for all Americans with AD/ADRD are estimated at 305 billion in 2020 \cite{Duan}. The high failure rate of the development of AD/ADRD drugs amplifies demographic and financial challenges. Given the increasing prevalence of the disease, finding innovative ways to develop effective drugs is an urgent need. Drug repurposing is a strategy for identifying new usages of approved or investigational drugs that are outside the scope of their original medical indications \cite{Ashburn}. There are majorly three computational methods for discovering drug repurposing evidence: the network-based methods, text mining and natural language processing (NLP) based approaches, as well as machine learning-based approaches \cite{Park2019}. Inspired by the fact that biologic entities in the same module of biological networks share similar characteristics, network-based approach aims to find several modules(subnetworks or cliques) using algorithms according to the topology structures of networks. NLP approaches usually includes processes of identifying biological entities and mining new knowledge from scientific literature. While machine learning-based approaches can apply different machine learning models such as logistic regression, support vector machine (SVM), random forest(RF), and deep learning (DL) to identify drug repurposing signals The computational drug repurposing strategy offers various advantages over developing entirely new drugs, including the possibilities to lower failure risks and risk of unknown side effects/complications, efficient utilization of development funds and shortened development timelines \cite{Pushpakom}. Developments in high-throughput screening technologies have catapulted computational drug repurposing to the forefront of attractive drug discovery approaches because the vast amounts of available data could potentially lead to new clues for drug repurposing that individual projects could not possibly reveal. 

Knowledge graphs can provide comprehensive and semantic representations for heterogeneous data, which has been successfully leveraged in many biomedical applications including drug repurposing \cite{bonner2021review}. For example, a few recent research focused on using knowledge graph-based approaches to drug repurposing for COVID-19 \cite{Zhang} \cite{Yan} \cite{AI-Saleem}. Sosa et al. applied knowledge graph embedding methods in drug repurposing for rare diseases \cite{Sosa}. Malas et al. leveraged the semantic properties of a knowledge graph to prioritize drug candidates for Autosomal Dominant Polycystic Kidney Disease (ADPKD) \cite{Malas}. However, to the best of our knowledge, knowledge graph-based approaches have rarely been applied in AD/ADRD drug repurposing. 

The objective of this paper is to study potential relations between Alzheimer’s diseases and dietary supplements, chemicals, and drugs using a knowledge graph-based approach. Studies have indicated that some drugs, chemicals or food supplements could be related to preventing or delaying neurodegeneration and cognitive decline \cite{Joseph}. However, further research is needed to better understand the back-end mechanisms and to reveal the potential interactions with clinical and pharmacokinetic factors. In this paper, we encode biomedical concepts and their rich relations into a knowledge graph through literature mining \cite{PubMed}. Literature Mining is a data mining technique that identifies the entities such as genes, diseases, and chemicals from literature, discovers global trends, and facilitates hypothesis generation based on existing knowledge. Literature mining enables researchers to study a massive amount of literature quickly and reveal hidden relations between entities that were hard to be discovered by manual analysis. More specifically, we introduce a biomedical knowledge graph that specifically focuses on AD/ADRD and discovers underlying relations between chemicals, drugs, dietary supplements and AD/ADRD. More details of how to construct the knowledge graph and how to leverage graph embedding methods to predict candidates with scoring will be described in the methods section. We also present several rankings of candidates and comparisons of different graph embedding algorithms.

\section*{Results and Discussion}

% after delete some duplicates: 791827= 649924 113031 28872 , still 112208 objects
% before BERT 2,811,329, 128177 objects
%   2,811,329 -  1672110
\subsection*{Knowledge graph construction}

There are 113,863,366 triples and 20,943,461 entities in total obtained from SemMedDB including 68 types of relations and 133 pairs of subject/object. After the rule-based filtering process described in the Preprocessing section, there are 2,811,329 triples left with a total of 128,177 subjects and objects. With further BERT-based filtering, 1,672,110 triples and 128,177 objects/subjects are left. After deduplicating triplets before training of graph embedding algorithms, there are 791,827 triples and 128,177 objects/subjects left.

\subsection*{Experimental settings}

All 791,827 triples are split into 649,924/113,031/28,872 as training/test/validation sets respectively. The split is done in a way that we can use data from 2019 to 2020 to validate our model and triples before 2019 as the training set and triples after 2020 as the test set. Table 1 shows the performance of three widely used graph completion methods that are trained on our knowledge graph: TransE is based on translational distance and DistMult and ComplEx are based on semantic information. We can see that the TransE model performs the best among all these graph embedding algorithms with a Mean Rank (MR) of 10.53 and a Hit Ratio of 10 (Hits@10) 0.58. We then use TransE model for the prediction of potential candidates. Specifically, the final model embeds nodes into a size of 250 with a learning rate of 0.01 with an L2 distance metric.

\begin{table}[h!]
\caption{Graph Embedding Algorithms Performance}
      \begin{tabular}{cccccc}
        \hline
           & MR  & MRR   & Hits@1 & Hits@3 & Hits@10\\ \hline
        TransE &10.53 &0.38 & 0.28 & 0.40 & 0.58\\
        DistMult &14.58 &0.23& 0.13 & 0.21  & 0.40\\
        ComplEx & 12.16  & 0.28 & 0.18& 0.29& 0.47         \\ \hline
      \end{tabular}
\end{table}

\subsection*{Prediction results}

We found that some potential candidates might be relevant to AD prevention and treatment. Based on the training data and our scoring function, we identified the top-ranked subjects that connect with AD-related concepts with predicates \textit{treat} or \textit{prevent}.  Tables 2, 4, and 6 show the top 10 entities according to their numbers of appearances for the \textit{drug, chemical}, and \textit{Dietary Supplement} categories respectively. Table 3,5 and 7 shows the top 10 ranked triples according to the candidate scores for the three categories. The triples with relevant evidence from PubMed with studies earlier than 1/1/2019 are marked in bold. The triples that only appeared in recently published studies after 1/1/2019 are marked in italic. The clinical drug and chemical categories were extracted from the Unified Medical Language System (UMLS) and we used the Integrated Dietary Supplement Knowledge Base (iDISK) \cite{Rizvi} as a reference for dietary supplements.

\subsubsection*{Clinical Drug}

 For the treatment relation, We were able to find evidence supporting seven out of ten entities (Table 2) and six out of ten triples (Table 3) through related literature and clinical trials for triples. All drugs appear in Table 4 appear in Table 2 while Table 2 has some extra drugs: Local corticosteroid, acyclovir, metronidazole, Cam, and Dexamethasone. Specifically, corticosteroids might become part of a multi-agent regimen for Alzheimer’s disease and also have applications for other neurodegenerative disorders \cite{Joseph2008}. Our model indicates that Valacyclovir, an antiviral medication might also have an effect in AD/ADRD prevention. While we did not find evidence that Acyclovir is directly related to AD/ADRD, a recent study shows that Valacyclovir Antiviral therapy could be used to reduce the risk of dementia \cite{Devanand}. A study demonstrated that antibiotic (ABX) cocktail-mediated perturbations (high dose kanamycin, gentamicin, colistin, metronidazole, vancomycin) of the gut microbiome in two independent transgenic lines leads to a reduction in A$\beta$ deposition in male mice and underlie the observed reductions in brain amyloidosis, which is the hallmark of Alzheimer's Disease.\cite{Hemraj2020}. 
 Tacrolimus \cite{Tacrolimus} has been in phase two clinical trial which investigates neurobiological effect in persons with MCI and dementia starting 12/1/2021. Early study also indicated that high doses of prednisolone have the effect of reducing amyloid reduction which resulted in some delay of the cognitive decline \cite{Alisky}\cite{Ricciarelli}. Propranolol \cite{Dobarro} has shown efficacy in reducing cognitive deficits in Alzheimer's transgenic mice. According to Joseph\cite{Joseph2008}, a short pulse of high dose intrathecal methylprednisolone, dexamethasone or triamcinalone will result in detectable slowing of Alzheimer’s disease. 
 
 As for the prevent relation, we found evidence that supports seven among ten triple predictions (Table 3) and all drugs in this table also appear in the Table 3. For example, a recent study in 2021 shows that Amifostine, which appears in our top 4 triple predictions, could mitigate cognitive injury induced by heavy-ion radiation \cite{Boutros}. Betaine could be a promising candidate for arresting Hcy-induced D-like pathological changes and memory deficits \cite{Chai2013}. Mazurek et al. show that Oxytocin could interfere with the formation of memory in experimental animals and contribut to memory disturbance associated with Alzheimer's disease \cite{Mazurek}.

\begin{table}[ht!]
\caption{Rankings For Drugs}
\resizebox{\columnwidth}{!}{
      \begin{tabular}{|cc|cc|}
        \hline
        Treat & Frequency &  Prevent & Frequency\\ \hline
          Imiquimod 50 mg/ml topical cream	&	4	&	 \it Amifostine	&	10	\\
             \bf Local corticosteroid injections	&	4	&	 \it Oral form acyclovir	&	6	\\
             \it Oral form acyclovir	&	4	&	Betadine ointment	&	5	\\
            \it Oral form metronidazole	&	4	&	 \bf  Betadine solution	&	5	\\
             \bf  Oral form prednisolone	&	4	&	 \it Oral form vancomycin	&	5	\\
             \bf  Oral form propranolol	&	4	&	 \bf  Oxytocin injectable solution	&	5	\\
            Pimecrolimus 10mg/g cream	&	4	&	Tenoxicam 20 mg	&	5	\\
             \it Topical form tacrolimus	&	4	&  \bf 	Betadine alcoholic	&	4	\\
            Cam, topical lotion	&	3	&	Bromfenac ophthalmic solution	&	4	\\
              \bf   Dexamethasone injection	&	3	&	Corticosteroids cream	&	4\\

      \hline
      \end{tabular}}
\end{table}

\begin{table}[ht!]
\caption{Rankings For Drug Triples}
\resizebox{\columnwidth}{!}
      {\begin{tabular}{|cc|cc|}
        \hline
        \multicolumn{2}{|c|}{Treat} & \multicolumn{2}{|c|}{Prevent}\\ \hline
          
        \it  Topical form tacrolimus	&	\it AD	&	\it Amifostine	&	\it  AD	\\									
        imiquimod Topical Cream	&	AD	&	\it Amifostine	&	\it AD, Late Onset	\\									
        \bf Oral form prednisolone	&	\bf AD	&	Betadine Ointment	&	AD	\\									
        Pimecrolimus cream	&	AD	&	\bf Betadine Solution	&	\bf AD	\\									
        \it Topical form tacrolimus	& \it AD, Early Onset	&	\it Oral form acyclovir	&	\it  AD\\									
        \bf Oral form propranolol	&	\bf AD	&	\it Amifostine	&	\it Familial AD	\\									
        imiquimod Topical Cream	& AD, Early Onset	&	\it Amifostine	&	\it AD, Early Onset	\\
        
        \it Topical form tacrolimus	&	\it Familial AD	&	\bf Oxytocin Injectable Solution	&	\bf AD\\									
        \bf Oral form prednisolone	&	\bf AD, Early Onset	&	tenoxicam 20 MG	&	AD\\									
        imiquimod  Topical Cream	&	Familial AD	&	\it Oral form acyclovir	&	\it AD, Late Onset	\\									
    
      \hline
      \end{tabular}}
\end{table}

\subsubsection*{Chemical}

 For the treat relationship prediction, we found supporting evidence for seven out of the top ten entities (Table 4) and eight out of the top ten triple predictions (Table 5). For the treat relations, Table 4 and Table 5 have some overlaps: Amifostine, Chlorhexidine, Amiloride, Etazolate, and licopyranocoumarin. As we discussed in the Drug section, Amifostine, which appears in our top 1 triple predictions, could mitigate cognitive injury induced by heavy-ion radiation \cite{Boutros}. Moreover, a study has shown that oral pathogens in some circumstances can approach the brain, potentially affecting memory and causing dementia \cite{Sureda}. Since chlorhexidine could be used to reduce Methicillin-resistant Staphylococcus aureus (MRSA) to improve oral health, it might be a potential candidate for the treatment of Alzheimer's Disease. Several studies mentioned the neuroprotective activity of Tetracycline and its derivatives \cite{Li} \cite{Bortolanza}.  Amiloride is a Na+/H+ exchangers (NHEs), which is proved to be associated with the development of mental disorders or Alzheimer's disease \cite{Verma}. In addition, we found in an earlier clinical trial that Etazolate was used to moderate AD \cite{Etazolate}. Licopyranocoumarin, as a compound from herbal medicine, was proved to have neuroprotective effect to Parkinson disease \cite{fuji}.
 
Dexrazoxane and Forskolin only appear in Table 4. A study in 2019 implies that Dexrazoxane may serve as an effective neuroprotectant to treat neurodegeneration and has potential clinical value in term of PD therapeutics\cite{MEI2019107758}. Forskolin shows neuroprotective effects in APP/PS1 Tg mice and may be a promising drug in the treatment of patients with AD\cite{Brice2017}. In addition, Tetracyline and proparglyamine only show up in Table 5. There are several studies mentioned that the neuroprotective activity of Tetracycline and its derivatives \cite{Li} \cite{Bortolanza}. Propargylamine was discussed on its beneficial effects and pro-survival/neurorescue inter-related activities relevant to Alzheimer's Disease in several studies \cite{do}\cite{Amit}. 

For prevention relation, we found six out of ten triples that are related to AD and all six corresponding chemicals also appear in Table 4. Recent studies show that antibiotic chemicals such as Fluoroquinolones, Amoxicillin, Clarithromycin, and Ampicillin can produce therapeutic effects to Alzheimer's Disease \cite{ou}\cite{Angelucci}.  Although we have not found that Cortisone has a direct effect on Alzheimer's Disease, common anti-inflammatory drugs do have some treatment effects \cite{Jaturapatporn}.  Earlier study has shown that allopurinol has treatment of aggressive behaviour in patients with dementia \cite{Lara2003}. In addition, Ceftriaxone(CEF) appears in Table 4. It significantly attenuated amyloid deposition and neuroinflammatory response and a study has confirmed the potential of CEF as a promising treatment against cognitive decline from the early stages of AD progression \cite{Tikhonova2021}.

\begin{table}[h!]
\caption{Rankings For Chemicals}
      \begin{tabular}{|cc|cc|}
        \hline
        Treat & Frequency &  Prevent & Frequency\\ \hline
         \bf Amifostine	&	4	& \bf	Amoxicillin	&	3	\\
       \it Chlorhexidine	&	4	& \bf	Cortisone	&	3	\\
         \bf Amiloride	&	2	& \it 	Fluoroquinolones	&	3	\\
        \it Dexrazoxane	&	2	&	Streptomycin	&	3	\\
         \bf Enoxaparin	&	2	&	\bf Allopurinol	&	2	\\
         \bf Etazolate	&	2	& \bf	Ampicillin	&	2	\\
         \bf  Forskolin	&	2	&	Aureomycin	&	2	\\
         \bf Licopyranocoumarin	&	2	&	\it  Ceftriaxone	&	2	\\
        Local anesthesia	&	2	&  \it 	Clarithromycin	&	2	\\
        M 40403	&	2	& Gabapentin	&	2	\\

      \hline
      \end{tabular}
\end{table}

\begin{table}[h!]
\caption{Rankings For Chemical Triples}
      \begin{tabular}{|cc|cc|}
        \hline
        \multicolumn{2}{|c|}{Treat} & \multicolumn{2}{|c|}{Prevent}\\ \hline
        \it  Chlorhexidine	&	\it  AD	&	\bf Cortisone & \bf  AD	\\
        
        \it  ritonavir	&AD	&\it Fluoroquinolones	&\it 	AD\\
        
        \bf Amifostine	&	\bf AD	&	Streptomycin	&	AD	\\
        
        \bf Tetracycline	&\bf 	AD	&\bf	Amoxicillin	&	\bf AD	\\
        
       \bf  Etazolate	&	\bf AD	&	Itraconazole	&	AD	\\
        
       \bf  licopyranocoumarin	& \bf AD	&	\bf Pentoxifylline	&\bf AD	\\

        \bf Amiloride	& \bf AD	&    \it 	Clarithromycin	&	   \it  AD	\\
        
        Seprafilm	&	AD	&	\bf Ampicillin	&	\bf AD	\\
        \bf propargylamine	&	\bf AD	&	Penicillins	&	AD	\\
        
        \it  Chlorhexidine	&\it AD, Late Onset	&	Aureomycin	&	AD	\\								
    
      \hline
      \end{tabular}
\end{table}

\subsubsection*{Dietary Supplement}

 Since there is little evidence that food can directly treat or prevent the Alzheimer's Disease, we focus on the triples with affect relationships. In the rank of the top 10 predictions of Table 7, we found dietary fiber (three times),  tea (three times), rice, and honey all have the possibility to reduce the risk of AD/ADRD and they also appear in Table 6. Dietary fiber has the potential that protects impact on brain A$\beta$ burden in older adults and the finding may assist in the development of dietary that prevent AD onset \cite{Fernando}. Moreover, according to \cite{kakutani}, green tea intake might reduce the risk of dementia and cognitive impairment. Another study shows that honey can be a rich source of cholinesterase inhibitors and therefore may play a role in AD treatment \cite{Baranowska}. Previous studies have also shown that dietary choline intake (e.g. eggs (egg yolk) and fruits) are associated with better outcomes on cognitive performance \cite{Maija}. Increasing dietary intake of minerals could also reduce the risk of dementia. For example, research found a link between potassium levels and diagnosis of cognitive impairment in Mexican-Americans. \cite{vintimilla2018}. In addition, one recent study indicates that highly water pressurized brown rice could ameliorate cognitive dysfunction and reduce the levels of amyloid-$\beta$, which is a major protein responsible for AD/ADRD \cite{Okuda}. Coffee drinking may be associated with a decreased risk of dementia/AD. This may be mediated by caffeine and/or other mechanisms like antioxidant capacity and increased insulin sensitivity.\cite{Eskelinen2010} Existing literature provides a reasonably strong scientific rationale to encourage testing whether ketamine (or its metabolites) has procognitive effects on Alzheimer's patients.\cite{Smalheiser2019}. Last but not least, based on the available literature, a nutraceutical formulation containing N-acetylcysteine among other compounds has shown some pro-cognitive benefits in Alzheimer's patients \cite{Hara2017}.

\begin{table}[h!]
\caption{Rankings For Dietary Supplements}
\begin{tabular}{|cc|}
\hline
Affect & Frequency \\
\hline
\bf Dietary fiber	&	8\\
\it Tea	&	8\\
\it Egg food product	&	7\\
\bf Electrolytes	&	7\\
\it Fruit	&	7\\
\it Honey	&	7\\
\bf Rice	&	7\\
\bf Coffee	&	6\\
\it  Ketamine	&	6\\
\bf Acetylcysteine	&	5\\

\hline
\end{tabular}
\end{table}

\begin{table}[h!]
\caption{Rankings For Dietary Supplement Triples}
\begin{tabular}{|cc|}
\hline
\multicolumn{2}{|c|}{Affect}\\
\hline
\it Tea	&	\it Familial AD \\
\bf Dietary Fiber	&	\bf Familial AD \\
\it Tea	&	\it AD 7 \\
\bf Rice	& \bf	Familial AD \\
\bf Dietary Fiber	&	\bf AD \\ 
\it Fruit	&	\it  Familial AD\\
 \it  Egg Food Product	& \it  Familial AD \\ 
\it Tea	& \it	AD, Late Onset \\ 
\it Honey	&	\it Familial AD \\ 
\bf Dietary Fiber	& \bf	AD 7\\
\hline
\end{tabular}
\end{table}

\section*{Conclusion}

In this study, we built a framework to construct and analyze a knowledge graph that links AD/ADRD-related biomedical knowledge from PubMed to facilitate drug repurposing. More specifically, we focused on identifying potentially new relationships between AD/ADRD and chemical, drug and food supplements respectively. Our analysis indicated that the pipeline can be used to identify biomedical concepts that are semantically close to each other as well as to reveal relationships between biomedical elements and diseases of interest. Linking sparse knowledge from fast-growing literature would be beneficial for existing knowledge/information retrieval, and may promote uncovering of new knowledge. This framework is flexible and can be used for other applications such as multi-omics applications, therapeutic discovery, and clinical decision support for neurodegenerative diseases as well as other diseases. The knowledge graph we constructed can facilitate data-driven knowledge discovery and new hypothesis generation. 

A breadth of possibilities exists to further improve this framework. First, our knowledge graph leveraged SemMedDB, an existing database that contains triples extracted from PubMed article. While we tried to improve the accuracy using a BERT-based approach, other NLP techniques could be implemented to further improve the accuracy of information extraction. Second, in addition to include knowledge extracted from literature, we could also incorporate triples from well-acknowledged biomedical databases to further enrich the knowledge graph. Third, we leveraged three state of the art knowledge graph embedding models in this research. In the future, we will investigate new strategies to extend embeddings to cope with sparse and unreliable data as well as multiple relationships. Last but not least, we only focused on the top 10 ranked triples for evaluation in this paper. We were able to identify supporting evidence for most of them, which indicates that our approach can inform reliable new knowledge. In addition, we only incorporate 2.8M triples for our knowledge graph due to computational resource limits, further investigation needs to be done on additional triples, which could potentially lead to new hypotheses for AD treatment and prevention.

\section*{Methods}

 We constructed a knowledge graph using biomedical concepts and relations extracted from  PubMed literature using NLP tools. The extracted triples were then further filtered based on statistics and NLP models. The rest of the subject-relation-object triples were used to build the knowledge graph. We then applied graph embedding algorithms to identify potential candidates for AD treatment and prevention. An overview of this is also described in Figure 1.

\subsection*{Data Collection and Relationship Extraction}
To construct the knowledge graph, we directly obtained triples from SemMedDB \cite{Kilicoglu}, which is a database of triples that are automatically extracted from the biomedical literature using Natural Language Processing (NLP) tools through SemRep \cite{Kilicoglu1}. Subject and object arguments are normalized to concepts defined in the UMLS with unique identifiers (CUIs). The triples are in the form of subject-predicate-object. 
%We used similar rules with \cite{Zhang} to exclude unrelated concepts and relations from these triples. Since the triples were extracted using SemRep, which has a reported precision rate of 0.69 {\color{cyan}\cite{Kilicoglu1}}, we used a BERT-based model, namely PubMedBERT \cite{pubmedbert} to further improve the accuracy. More detail of this will be discussed later in "Calibration using PubMedBERT" part.%

\begin{figure}[h!]
     \begin{center}
     \includegraphics[width=6cm]{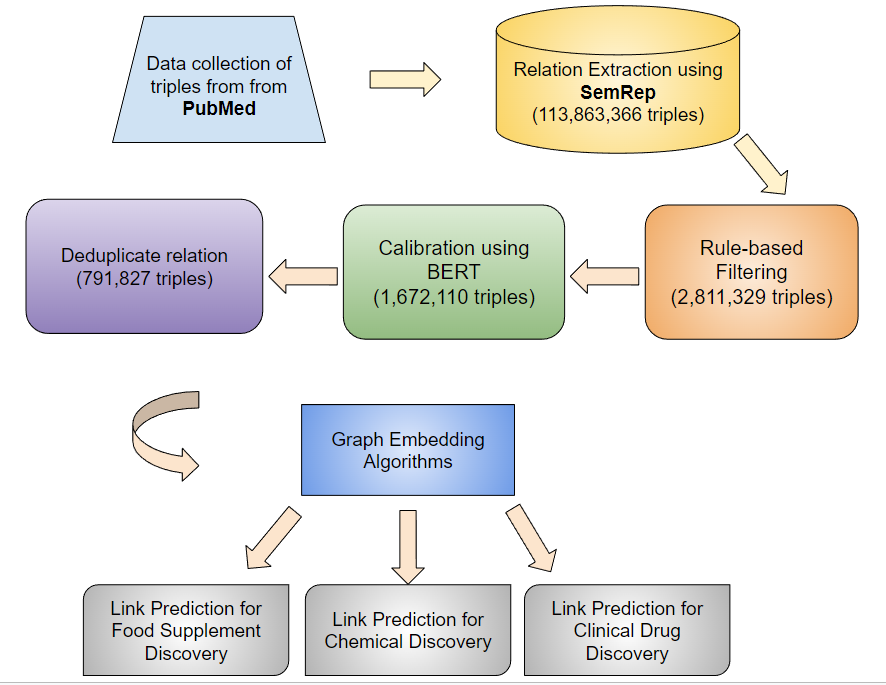}
    \end{center}
\caption{\csentence{General Pipeline: The biological concepts in PubMed literature was extracted using NLP tools and was built into a knowledge graph using Subject-relation-object triples. Graph embedding algorithms were used to find potential candidates and complete the knowledge graph. Number of triples left are shown in each step.}}
  \end{figure}

\subsection*{Rule-based Filtering}

The original data directly obtained from the SemMedDB contained a large number of triples, but not all of them are useful for finding candidates for AD/ADRD treatment/prevention. We applied rules that are similar to \cite{Zhang} to exclude unrelated subject/object and predicate types. More specifically,  we eliminated triples involving generic biomedical concepts such as Activities $\&$ Behaviors, Concepts $\&$ Ideas, Objects, Occupations, Organizations, and Phenomena. The rest of the triples were eliminated based on their degree of centrality ($A_{in} , A_{out}$) and $G^{2}$ score that indicates the strength of association between a subject and an object. Specifically, the degree centrality($A_{in} , A_{out}$) was calculated with the adjacency matrix M as: 

\begin{ceqn}
\begin{align}
            A_{in}  =  \sum_{j=1}^{n} M_{ji}    \  and  \ A_{out} =  \sum_{j=1}^{n} M_{ij}
\end{align}
\end{ceqn}

And the $G^{2}$ score is calculated from the statistical relation between two contingency tables: Observation table and Expectation table.\cite{McInnes}

\begin{ceqn}
\begin{align}
            G^{2}  =  2 \sum_{i,j,k}^{} O_{ijk} *  log (\frac{O_{ijk}}{E_{ijk}})
\end{align}

\end{ceqn}
where $O_{ijk}$ represents the items in the observation table and 
\begin{ceqn}
\begin{align}
  E_{ijk} = \frac{\sum_{i}^{} O_{ijk} \sum_{j}^{} O_{ijk} \sum_{i}^{} O_{ijk}}{(\sum_{}^{} O_{ijk})^{2}}
\end{align}

\end{ceqn}
represents the items in the expectation table. 

At last, these three scores were normalized to [0, 1] and summed up into a final score. To keep the knowledge graph in a reasonable size that the graph embedding algorithms could handle, we only kept about 2.5M triples. In order to ensure that AD-related triples are included in the knowledge graph, we kept all triples that are related to Alzheimer's Diseases terms in the UMLS during triple elimination using the above criteria. Table 9 in the additional file section summarizes the AD-related UMLS concepts we kept in this process. At last, we have 2.8M triples left in our knowledge graph.

\subsection*{Calibration using PubMedBERT}

We leveraged about 6,000 annotations from a previous study \cite{Rizvi} and used them as the training data for the PubMedBERT fine-tuning. These annotations were manually labeled with 1 or 0, where 1 indicates that the triples and their relationships do exist and are correct (triples labeled with 1); and 0 means that the triples do not exist or are incorrect (triples labeled with 0). PubMedBERT took the text input of subject, object, predicate type as well as the sentence that these were extracted from. The model obtained an F-1 score of 0.82, Recall of 0.91 and Precision of 0.75 on the validation set; and  F-1 score of 0.83, Recall of 0.89 and Precision of 0.78 on these annotations. 
% Best test results:
% Loss: 0.420571	F1: 0.8279
% Recall: 0.8879	Precision: 0.7755
% Validation Results:
% 		Loss: 0.646441	F1: 0.8227
% 			Recall: 0.9062	Precision: 0.7532
% New best f1! Saving weights...

\subsection*{Graph Embedding Algorithms}

% https://aclanthology.org/D19-1246.pdf

Knowledge graph embedding is a promising approach to graph completion tasks \cite{Lin}. It embeds entities and relations into vector space to evaluate the probability that a given triplet (h,r,t) is true through a scoring function. We leveraged three popular knowledge graph embedding methods, TransE, DistMult and ComplEx for our knowledge graph completion task. To train this knowledge graph, these three models do negative sampling by corrupting triplets (h,r,t) to either form (h',r,r) or (h,r,t'), where h' and t' are the negative samples. Therefore, if y=$\pm1$ is the label for positive and negative triplets and f is the scoring function, then the logistic loss is computed as according to \cite{DGL-KE}:

\begin{ceqn}
\begin{align}
            \sum_{(h,r,t) \in D^{+} \cup D^{-}}^{n} log(1+e^{-y*f(h,r,t)})
\end{align}
\end{ceqn}

\section*{Tables}
\begin{table}[h!]
\caption{Scoring Function of Graph Embedding Algorithms}
      \begin{tabular}{ccc}
        \hline
        Model & Scoring Type  & Scoring Function \\ \hline
        TransE & Distance Function & -$\lVert \mathbf{h+r-t} \rVert$ \\
        ComplEx & Bilinear Function &  $Re(h^{T}(diag(r))\Bar{t})$\\
        DistMult & Bilinear Function &  $h^{T} Re(diag(r))t$\\
        \hline
      \end{tabular}
\end{table}

\subsubsection*{TransE}

TransE \cite{Bordes} is one of the earliest translational distance models. The model projects head, tail and relations into the same space where the relation is interpreted as a translation vector r so that the head and tail can be connected by relations with low error. And the score function is the negative of the distance of this error as shown in Table 8. TransE does have disadvantages in dealing with 1-to-N, N-to-1, and N-to-N relations. For example, if Alzheimer's Disease could be affected by different food supplements, then TransE model might learn similar results for all these food supplements.

\subsubsection*{DistMult}

Semantic matching models like DistMult\cite{yang} use similarity-based scoring functions that associate each entity with a vector to capture its latent semantics. In this model, each relation is represented as a diagonal matrix which models pairwise interactions between latent factors by a bilinear function as shown in Table 8.

\subsubsection*{ComplEx}

Since the scoring function of DistMult is symmetric in terms of h and t, the function cannot handle asymmetric relationships. Complex Embeddings (ComplEx) \cite{trouillon2016complex} introduces complex-valued embeddings to solve this problems. Specifically, the scoring function can be expanded as:

\begin{ceqn}
\begin{align}
 Re(h^{T}(diag(r))\Bar{t} = Re(\sum_{i=0}^{d-1} [r]_{i} [r]_{i} [\bar{t}]_{i})
\end{align}
\end{ceqn}

\subsection*{Candidates scoring for repurposing}

We focused on three kinds of predictions for the candidate selection in this research: dietary supplements candidates, chemical candidates, and clinical drug candidates. The clinical drug and chemical categories were extracted from the UMLS and we used  the iDISK \cite{Rizvi} as a reference for dietary supplements. For each type of candidates, the model iterates over all possible triples, i.e. ($h_{i}$,$r_{i}$,$t_{k}$), and $h_{i}$ $\in$ {all nodes for particular type of candidates} ,$r_{j}$ $\in$ {all relations}, and $t_{k}$ $\in$ {all nodes related to Alzheimer's Disease}. In knowledge graph embedding–based approaches, the scoring function $\phi$(h, r, t) is defined in terms of the embeddings of entities and relations; i.e., h, r, and t are embedded into vector space, and $\phi$ is defined in terms of operations or scoring functions over these objects. They all project the node and entities to lower-dimensional embeddings but with different scoring functions. TransE simply uses the distance between the embeddings of the head, sum with the relation embedding and tail as the scoring function, while DistMult and ComplEx use bilinear map to define scoring functions. For drugs and chemicals, we used two types of relations (i.e., treat and prevent) for prediction in this paper since the focus of the paper is drug repurposing. For dietary supplements, on the other hand, we focus on the “affect” relationship since it might be relatively challenging to detect top-ranked direct relationships between dietary supplements and AD treatment/prevention.

\subsection*{Evaluation for drug repurposing}
We leveraged the time-slicing technique that is commonly used in literature mining \cite{HENRY201720} to evaluate our triple prediction approach. We trained all three models using data before 1/1/2019 to see whether we can predict triples that were first published after this date. 

%%%%%%%%%%%%%%%%%%%%%%%%%%%%%%%%%%%%%%%%%%%%%%
%%                                          %%
%% Backmatter begins here                   %%
%%                                          %%
%%%%%%%%%%%%%%%%%%%%%%%%%%%%%%%%%%%%%%%%%%%%%%

\begin{backmatter}

\section*{Acknowledgements}
None
 
\section*{Funding}
Publication costs are funded by the National Institute of the Aging of NIH under Award Number RF1AG072799. This research was supported by NIH grants under Award Numbers RF1AG072799, R01AI130460, and R01AT009457.

\section*{Abbreviations}
AD - Alzheimer's Disease

MR - Mean Rank

Hits@1/3/10 - Hit ratio at one/three/ten

ADRD - Alzheimer’s disease and related dementias

NLP - natural language processing

SVM - support vector machine

RF - random forest 

DL - deep learning

ADPKD - Autosomal Dominant Polycystic Kidney Disease

MRR - Mean Reciprocal Rank

iDISK - Integrated Dietary Supplement Knowledge Base

ABX - Antibiotic

MRSA - Methicillin-resistant Staphylococcus aureus

NHEs - Na+/H+ exchangers

CEF - Ceftriaxone

CUIs - UMLS concept unique identifiers

ComplEx - Complex Embeddings

UMLS - Unified Medical Language System

\section*{Availability of data and materials}
SemMedDB: \url{https://lhncbc.nlm.nih.gov/ii/tools/SemRep_SemMedDB_SKR.html}
UMLS: \url{https://www.nlm.nih.gov/research/umls/index.html}
iDISK:\url{https://academic.oup.com/jamia/article/27/4/539/5740032?login=true}

\section*{Ethics approval and consent to participate}
Not applicable

\section*{Competing interests}
The authors declare that they have no competing interests.
\section*{Consent for publication}
Not applicable

\section*{Author's contributions}
 CT conceived the research project. YN, JD and CT designed the pipeline and method. YN implemented the deep learning model of the study and prepared the manuscript. JF and FL conducted the result interpretation. XH and LB prepared the data and proceed the pipeline. RZ, YC, and YZ provided expertise and suggestions on data filtering and model design especially for the dietary supplement data. All authors proofread the paper and provided valuable suggestions. All the authors have read and approved the final manuscript.

%%  After submission of the .TEX file,                     %%
%%  you will be prompted to submit your .BBL file.         %%

% if your bibliography is in bibtex format, use those commands:
\bibliographystyle{bmc-mathphys} % Style BST file (bmc-mathphys, vancouver, spbasic).
\bibliography{bmc_article}      % Bibliography file (usually '*.bib' )


\begin{thebibliography}{58}
\ifx \bisbn   \undefined \def \bisbn  #1{ISBN #1}\fi
\ifx \binits  \undefined \def \binits#1{#1}\fi
\ifx \bauthor  \undefined \def \bauthor#1{#1}\fi
\ifx \batitle  \undefined \def \batitle#1{#1}\fi
\ifx \bjtitle  \undefined \def \bjtitle#1{#1}\fi
\ifx \bvolume  \undefined \def \bvolume#1{\textbf{#1}}\fi
\ifx \byear  \undefined \def \byear#1{#1}\fi
\ifx \bissue  \undefined \def \bissue#1{#1}\fi
\ifx \bfpage  \undefined \def \bfpage#1{#1}\fi
\ifx \blpage  \undefined \def \blpage #1{#1}\fi
\ifx \burl  \undefined \def \burl#1{\textsf{#1}}\fi
\ifx \doiurl  \undefined \def \doiurl#1{\textsf{#1}}\fi
\ifx \betal  \undefined \def \betal{\textit{et al.}}\fi
\ifx \binstitute  \undefined \def \binstitute#1{#1}\fi
\ifx \binstitutionaled  \undefined \def \binstitutionaled#1{#1}\fi
\ifx \bctitle  \undefined \def \bctitle#1{#1}\fi
\ifx \beditor  \undefined \def \beditor#1{#1}\fi
\ifx \bpublisher  \undefined \def \bpublisher#1{#1}\fi
\ifx \bbtitle  \undefined \def \bbtitle#1{#1}\fi
\ifx \bedition  \undefined \def \bedition#1{#1}\fi
\ifx \bseriesno  \undefined \def \bseriesno#1{#1}\fi
\ifx \blocation  \undefined \def \blocation#1{#1}\fi
\ifx \bsertitle  \undefined \def \bsertitle#1{#1}\fi
\ifx \bsnm \undefined \def \bsnm#1{#1}\fi
\ifx \bsuffix \undefined \def \bsuffix#1{#1}\fi
\ifx \bparticle \undefined \def \bparticle#1{#1}\fi
\ifx \barticle \undefined \def \barticle#1{#1}\fi
\ifx \bconfdate \undefined \def \bconfdate #1{#1}\fi
\ifx \botherref \undefined \def \botherref #1{#1}\fi
\ifx \url \undefined \def \url#1{\textsf{#1}}\fi
\ifx \bchapter \undefined \def \bchapter#1{#1}\fi
\ifx \bbook \undefined \def \bbook#1{#1}\fi
\ifx \bcomment \undefined \def \bcomment#1{#1}\fi
\ifx \oauthor \undefined \def \oauthor#1{#1}\fi
\ifx \citeauthoryear \undefined \def \citeauthoryear#1{#1}\fi
\ifx \endbibitem  \undefined \def \endbibitem {}\fi
\ifx \bconflocation  \undefined \def \bconflocation#1{#1}\fi
\ifx \arxivurl  \undefined \def \arxivurl#1{\textsf{#1}}\fi
\csname PreBibitemsHook\endcsname

%%% 1
\bibitem{neurodegenerative}
\begin{botherref}
Neurodegenerative diseases Latest research and news.
Accessed on 01.10.2022.
\url{https://www.nature.com/subjects/neurodegenerative-diseases}
\end{botherref}
\endbibitem

%%% 2
\bibitem{Moya-Alvarado}
\begin{botherref}
\oauthor{\bsnm{Moya-Alvarado}, \binits{G.}},
\oauthor{\bsnm{Gershoni-Emek}, \binits{N.}},
\oauthor{\bsnm{Perlson}, \binits{E.}},
\oauthor{\bsnm{Bronfman}, \binits{F.}}:
Neurodegeneration and alzheimer's disease (ad). what can proteomics tell us
  about the alzheimer's brain?.
Mol Cell Proteomics.
\textbf{15(2)}
(2016)
\end{botherref}
\endbibitem

%%% 3
\bibitem{Duan}
\begin{botherref}
\oauthor{\bsnm{Duan}, \binits{R.}},
\oauthor{\bsnm{Boland}, \binits{M.}},
\oauthor{\bsnm{Liu}, \binits{Z.}},
\oauthor{\bsnm{Liu}, \binits{Y.}},
\oauthor{\bsnm{Chang}, \binits{H.}},
\oauthor{\bsnm{Xu}, \binits{H.}},
\oauthor{\bsnm{Chu}, \binits{H.}},
\oauthor{\bsnm{Schmid}, \binits{C.}},
\oauthor{\bsnm{Forrest}, \binits{C.}},
\oauthor{\bsnm{Holmes}, \binits{J.}},
\oauthor{\bsnm{Schuemie}, \binits{M.}},
\oauthor{\bsnm{Berlin}, \binits{J.}},
\oauthor{\bsnm{Moore}, \binits{J.}},
\oauthor{\bsnm{Chen}, \binits{Y.}}:
Learning from electronic health records across multiple sites: A
  communication-efficient and privacy-preserving distributed algorithm.
J Am Med Inform Assoc.
\textbf{27(3)}
(2020)
\end{botherref}
\endbibitem

%%% 4
\bibitem{Ashburn}
\begin{botherref}
\oauthor{\bsnm{Ashburn}, \binits{T.}},
\oauthor{\bsnm{Thor}, \binits{K.}}:
Drug repositioning: identifying and developing new uses for existing drugs.
Nat Rev Drug Discov
\textbf{3}
(2004)
\end{botherref}
\endbibitem

%%% 5
\bibitem{Park2019}
\begin{barticle}
\bauthor{\bsnm{Kyungsoo}, \binits{P.}}:
\batitle{A review of computational drug repurposing}.
\bjtitle{tcp}
\bvolume{27}(\bissue{2}),
\bfpage{59}--\blpage{63}
(\byear{2019}).
doi:\doiurl{10.12793/tcp.2019.27.2.59}.
\arxivurl{http://www.e-sciencecentral.org/articles/?scid=1130297}
\end{barticle}
\endbibitem

%%% 6
\bibitem{Pushpakom}
\begin{botherref}
\oauthor{\bsnm{Pushpakom}, \binits{S.}},
\oauthor{\bsnm{Iorio}, \binits{F.}},
\oauthor{\bsnm{Eyers}, \binits{P.}},
\oauthor{\bsnm{Escott}, \binits{K.}},
\oauthor{\bsnm{Hopper}, \binits{S.}},
\oauthor{\bsnm{Wells}, \binits{A.}},
\oauthor{\bsnm{Doig}, \binits{A.}},
\oauthor{\bsnm{Guilliams}, \binits{T.}},
\oauthor{\bsnm{Latimer}, \binits{J.}},
\oauthor{\bsnm{McNamee}, \binits{C.}},
\oauthor{\bsnm{Norris}, \binits{A.}},
\oauthor{\bsnm{Sanseau}, \binits{P.}},
\oauthor{\bsnm{Cavalla}, \binits{D.}},
\oauthor{\bsnm{Pirmohamed}, \binits{M.}}:
Drug repurposing: progress, challenges and recommendations.
Nat Rev Drug Discov.
\textbf{18(1)}
(2019)
\end{botherref}
\endbibitem

%%% 7
\bibitem{bonner2021review}
\begin{botherref}
\oauthor{\bsnm{Bonner}, \binits{S.}},
\oauthor{\bsnm{Barrett}, \binits{I.}},
\oauthor{\bsnm{Ye}, \binits{C.}},
\oauthor{\bsnm{Swiers}, \binits{R.}},
\oauthor{\bsnm{Engkvist}, \binits{O.}},
\oauthor{\bsnm{Bender}, \binits{A.}},
\oauthor{\bsnm{Hoyt}, \binits{C.}},
\oauthor{\bsnm{Hamilton}, \binits{W.}}:
A review of biomedical datasets relating to drug discovery: A knowledge graph
  perspective.
arXiv preprint arXiv:2102.10062
(2021)
\end{botherref}
\endbibitem

%%% 8
\bibitem{Zhang}
\begin{botherref}
\oauthor{\bsnm{Zhang}, \binits{R.}},
\oauthor{\bsnm{Hristovski}, \binits{D.}},
\oauthor{\bsnm{Schutte}, \binits{D.}},
\oauthor{\bsnm{Kastrin}, \binits{A.}},
\oauthor{\bsnm{Fiszman}, \binits{M.}},
\oauthor{\bsnm{Kilicoglu}, \binits{H.}}:
Drug repurposing for covid-19 via knowledge graph completion.
Journal of biomedical informatics
\textbf{115 p.103696.}
(2021)
\end{botherref}
\endbibitem

%%% 9
\bibitem{Yan}
\begin{botherref}
\oauthor{\bsnm{Yan}, \binits{V.}},
\oauthor{\bsnm{Li}, \binits{X.}},
\oauthor{\bsnm{Ye}, \binits{X.}},
\oauthor{\bsnm{Ou}, \binits{M.}},
\oauthor{\bsnm{Luo}, \binits{R.}},
\oauthor{\bsnm{Zhang}, \binits{Q.}},
\oauthor{\bsnm{Tang}, \binits{B.}},
\oauthor{\bsnm{BJ}, \binits{C.}},
\oauthor{\bsnm{I}, \binits{H.}},
\oauthor{\bsnm{Siu}, \binits{C.}},
\oauthor{\bsnm{ICK}, \binits{W.}},
\oauthor{\bsnm{RCK}, \binits{C.}},
\oauthor{\bsnm{EW}, \binits{C.}}:
Drug repurposing for the treatment of covid-19: A knowledge graph approach.
Epub ahead of print. Erratum in: Adv Ther (Weinh).
\textbf{4(10):2100179}
(2021)
\end{botherref}
\endbibitem

%%% 10
\bibitem{AI-Saleem}
\begin{barticle}
\bauthor{\bsnm{Al-Saleem}, \binits{J.}},
\bauthor{\bsnm{Granet}, \binits{R.}},
\bauthor{\bsnm{Ramakrishnan}, \binits{S.}},
\bauthor{\bsnm{Ciancetta}, \binits{N.}},
\bauthor{\bsnm{Saveson}, \binits{C.}},
\bauthor{\bsnm{Gessner}, \binits{C.}},
\bauthor{\bsnm{Zhou}, \binits{Q.}}:
\batitle{Knowledge graph-based approaches to drug repurposing for covid-19}.
\bjtitle{Journal of Chemical Information and Modeling}
\bvolume{61}(\bissue{8}),
\bfpage{4058}--\blpage{4067}
(\byear{2021}).
doi:\doiurl{10.1021/acs.jcim.1c00642}.
\bcomment{PMID: 34297570}
\end{barticle}
\endbibitem

%%% 11
\bibitem{Sosa}
\begin{botherref}
\oauthor{\bsnm{Sosa}, \binits{D.}},
\oauthor{\bsnm{Derry}, \binits{A.}},
\oauthor{\bsnm{Guo}, \binits{M.}},
\oauthor{\bsnm{Wei}, \binits{E.}},
\oauthor{\bsnm{Brinton}, \binits{C.}},
\oauthor{\bsnm{Altman}, \binits{R.}}:
A literature-based knowledge graph embedding method for identifying drug
  repurposing opportunities in rare diseases.
Pac Symp Biocomput
\textbf{25}
(2020)
\end{botherref}
\endbibitem

%%% 12
\bibitem{Malas}
\begin{botherref}
\oauthor{\bsnm{Malas}, \binits{T.}},
\oauthor{\bsnm{Vlietstra}, \binits{W.}},
\oauthor{\bsnm{Kudrin}, \binits{R.}},
\oauthor{\bsnm{Starikov}, \binits{S.}},
\oauthor{\bsnm{Charrout}, \binits{M.}},
\oauthor{\bsnm{Roos}, \binits{M.}},
\oauthor{\bsnm{Peters}, \binits{D.}},
\oauthor{\bsnm{Kors}, \binits{J.}},
\oauthor{\bsnm{Vos}, \binits{R.}},
\oauthor{\bsnm{PAC}, \binits{H.}},
\oauthor{\bsnm{Mulligen}, \binits{E.}},
\oauthor{\bsnm{Hettne}, \binits{K.}}:
Drug prioritization using the semantic properties of a knowledge graph.
Sci Rep.
\textbf{9(1):6281}
(2019)
\end{botherref}
\endbibitem

%%% 13
\bibitem{Joseph}
\begin{botherref}
\oauthor{\bsnm{Joseph}, \binits{J.}},
\oauthor{\bsnm{Cole}, \binits{G.}},
\oauthor{\bsnm{Head}, \binits{E.}},
\oauthor{\bsnm{D}, \binits{I.}}:
Nutrition, brain aging, and neurodegeneration.
Journal of Neuroscience
\textbf{29}
(2009)
\end{botherref}
\endbibitem

%%% 14
\bibitem{PubMed}
\begin{botherref}
PubMed.
Accessed on 01.10.2022.
\url{https://pubmed.ncbi.nlm.nih.gov/}
\end{botherref}
\endbibitem

%%% 15
\bibitem{Rizvi}
\begin{barticle}
\bauthor{\bsnm{Rizvi}, \binits{R.}},
\bauthor{\bsnm{Vasilakes}, \binits{J.}},
\bauthor{\bsnm{Adam}, \binits{T.}},
\bauthor{\bsnm{Melton}, \binits{G.}},
\bauthor{\bsnm{Bishop}, \binits{J.}},
\bauthor{\bsnm{Bian}, \binits{J.}},
\bauthor{\bsnm{Tao}, \binits{C.}},
\bauthor{\bsnm{Zhang}, \binits{R.}}:
\batitle{{iDISK: the integrated DIetary Supplements Knowledge base}}.
\bjtitle{Journal of the American Medical Informatics Association}
\bvolume{27}(\bissue{4}),
\bfpage{539}--\blpage{548}
(\byear{2020})
\end{barticle}
\endbibitem

%%% 16
\bibitem{Joseph2008}
\begin{botherref}
\oauthor{\bsnm{Alisky}, \binits{J.M.}}:
Intrathecal corticosteroids might slow alzheimer's disease progression.
Neuropsychiatric disease and treatment.
\textbf{45}
(2008)
\end{botherref}
\endbibitem

%%% 17
\bibitem{Devanand}
\begin{botherref}
\oauthor{\bsnm{Devanand}, \binits{D.}},
\oauthor{\bsnm{Andrews}, \binits{H.}},
\oauthor{\bsnm{Kreisl}, \binits{W.}},
\oauthor{\bsnm{Razlighi}, \binits{Q.}},
\oauthor{\bsnm{Gershon}, \binits{A.}},
\oauthor{\bsnm{Stern}, \binits{Y.}},
\oauthor{\bsnm{Mintz}, \binits{A.}},
\oauthor{\bsnm{Wisniewski}, \binits{T.}},
\oauthor{\bsnm{Acosta}, \binits{E.}},
\oauthor{\bsnm{Pollina}, \binits{J.}},
\oauthor{\bsnm{Katsikoumbas}, \binits{M.}},
\oauthor{\bsnm{Bell}, \binits{K.}},
\oauthor{\bsnm{Pelton}, \binits{G.}},
\oauthor{\bsnm{Deliyannides}, \binits{D.}},
\oauthor{\bsnm{Prasad}, \binits{K.}},
\oauthor{\bsnm{Huey}, \binits{E.}}:
Antiviral therapy: Valacyclovir treatment of alzheimer's disease (valad) trial:
  protocol for a randomised, double-blind,placebo-controlled, treatment trial.
BMJ Open
\textbf{10(2):e0321112}
(2020)
\end{botherref}
\endbibitem

%%% 18
\bibitem{Hemraj2020}
\begin{botherref}
\oauthor{\bsnm{Dodiya}, \binits{H.}},
\oauthor{\bsnm{Frith}, \binits{M.}},
\oauthor{\bsnm{Sidebottom}, \binits{A.}},
\oauthor{\bsnm{Cao}, \binits{Y.}},
\oauthor{\bsnm{Koval}, \binits{J.}},
\oauthor{\bsnm{Chang}, \binits{E.}},
\oauthor{\bsnm{Sisodia}, \binits{S.}}:
Synergistic depletion of gut microbial consortia, but not individual
  antibiotics, reduces amyloidosis in appps1-21 alzheimer's transgenic mice.
Scientific reports.
(2020)
\end{botherref}
\endbibitem

%%% 19
\bibitem{Tacrolimus}
\begin{botherref}
A Pilot Open Labeled Study of Tacrolimus in Alzheimer's Disease.
Accessed on 01.10.2022.
\url{https://clinicaltrials.gov/ct2/show/results/NCT04263519}
\end{botherref}
\endbibitem

%%% 20
\bibitem{Alisky}
\begin{botherref}
\oauthor{\bsnm{Alisky}, \binits{J.}}:
Intrathecal corticosteroids might slow alzheimer's disease progression.
Neuropsychiatr Dis Treat
\textbf{4(5)}
(2008)
\end{botherref}
\endbibitem

%%% 21
\bibitem{Ricciarelli}
\begin{botherref}
\oauthor{\bsnm{Ricciarelli}, \binits{R.}},
\oauthor{\bsnm{Fedele}, \binits{E.}}:
The amyloid cascade hypothesis in alzheimer's disease: It's time to change our
  mind.
Curr Neuropharmacol
\textbf{15(6)}
(2017)
\end{botherref}
\endbibitem

%%% 22
\bibitem{Dobarro}
\begin{botherref}
\oauthor{\bsnm{Dobarro}, \binits{M.}},
\oauthor{\bsnm{Gerenu}, \binits{G.}},
\oauthor{\bsnm{Ramírez}, \binits{M.}}:
Propranolol reduces cognitive deficits, amyloid and tau pathology in
  alzheimer's transgenic mice.
Int J Neuropsychopharmacol
\textbf{16(10)}
(2013)
\end{botherref}
\endbibitem

%%% 23
\bibitem{Boutros}
\begin{botherref}
\oauthor{\bsnm{Boutros}, \binits{S.}},
\oauthor{\bsnm{Zimmerman}, \binits{B.}},
\oauthor{\bsnm{Nagy}, \binits{S.}},
\oauthor{\bsnm{Lee}, \binits{J.}},
\oauthor{\bsnm{Perez}, \binits{R.}},
\oauthor{\bsnm{Raber}, \binits{J.}}:
Amifostine (wr-2721) mitigates cognitive injury induced by heavy ion radiation
  in male mice and alters behavior and brain connectivity.
Frontiers in Physiology
\textbf{12}
(2021).
doi:\doiurl{10.3389/fphys.2021.770502}
\end{botherref}
\endbibitem

%%% 24
\bibitem{Chai2013}
\begin{botherref}
\oauthor{\bsnm{Chai}, \binits{G.-S.}},
\oauthor{\bsnm{Jiang}, \binits{X.}},
\oauthor{\bsnm{Ni}, \binits{Z.-F.}},
\oauthor{\bsnm{Ma}, \binits{Z.-W.}},
\oauthor{\bsnm{Xie}, \binits{A.-J.}},
\oauthor{\bsnm{Cheng}, \binits{X.-S.}},
\oauthor{\bsnm{Wang}, \binits{Q.}},
\oauthor{\bsnm{Wang}, \binits{J.-Z.}},
\oauthor{\bsnm{Liu}, \binits{G.-P.}}:
Betaine attenuates alzheimer-like pathological changes and memory deficits
  induced by homocysteine.
J Neurochem.
\textbf{(3):388-96.}
(2013, Feb)
\end{botherref}
\endbibitem

%%% 25
\bibitem{Mazurek}
\begin{botherref}
\oauthor{\bsnm{Mazurek}, \binits{M.}},
\oauthor{\bsnm{Beal}, \binits{M.}},
\oauthor{\bsnm{Bird}, \binits{E.}},
\oauthor{\bsnm{Martin}, \binits{J.}}:
Oxytocin in alzheimer's disease: postmortem brain levels.
Neurology
\textbf{37(6)}
(1987)
\end{botherref}
\endbibitem

%%% 26
\bibitem{Sureda}
\begin{botherref}
\oauthor{\bsnm{Sureda}, \binits{A.}},
\oauthor{\bsnm{Daglia}, \binits{M.}},
\oauthor{\bsnm{Castilla}, \binits{S.}},
\oauthor{\bsnm{Sanadgol}, \binits{N.}},
\oauthor{\bsnm{Nabavi}, \binits{S.}},
\oauthor{\bsnm{Khan}, \binits{H.}},
\oauthor{\bsnm{Belwal}, \binits{T.}},
\oauthor{\bsnm{Jeandet}, \binits{P.}},
\oauthor{\bsnm{Marchese}, \binits{A.}},
\oauthor{\bsnm{Pistollato}, \binits{F.}},
\oauthor{\bsnm{Forbes-Hernandez}, \binits{T.}},
\oauthor{\bsnm{Battino}, \binits{M.}},
\oauthor{\bsnm{Berindan-Neagoe}, \binits{I.}},
\oauthor{\bsnm{G}, \binits{D.}},
\oauthor{\bsnm{Nabavi}, \binits{S.}}:
Oral microbiota and alzheimer's disease: Do all roads lead to rome?.
Pharmacol Res
\textbf{151:104582}
(2020)
\end{botherref}
\endbibitem

%%% 27
\bibitem{Li}
\begin{botherref}
\oauthor{\bsnm{Li}, \binits{C.}},
\oauthor{\bsnm{Yuan}, \binits{K.}},
\oauthor{\bsnm{Schluesener}, \binits{H.}}:
Impact of minocycline on neurodegenerative diseases in rodents: a
  meta-analysis.
Rev Neurosci.
\textbf{24(5)}
(2013)
\end{botherref}
\endbibitem

%%% 28
\bibitem{Bortolanza}
\begin{botherref}
\oauthor{\bsnm{Bortolanza}, \binits{M.}},
\oauthor{\bsnm{Nascimento}, \binits{G.}},
\oauthor{\bsnm{Socias}, \binits{S.}},
\oauthor{\bsnm{Ploper}, \binits{D.}},
\oauthor{\bsnm{Chehín}, \binits{R.}},
\oauthor{\bsnm{Raisman-Vozari}, \binits{R.}},
\oauthor{\bsnm{Del-Bel}, \binits{E.}}:
Tetracycline repurposing in neurodegeneration: focus on parkinson's disease.
J Neural Transm (Vienna).
\textbf{125(10)}
(2018)
\end{botherref}
\endbibitem

%%% 29
\bibitem{Verma}
\begin{botherref}
\oauthor{\bsnm{Verma}, \binits{V.}},
\oauthor{\bsnm{Bali}, \binits{A.}},
\oauthor{\bsnm{Singh}, \binits{N.}},
\oauthor{\bsnm{Jaggi}, \binits{A.}}:
Implications of sodium hydrogen exchangers in various brain diseases.
J Basic Clin Physiol Pharmacol.
\textbf{26(5)}
(2015)
\end{botherref}
\endbibitem

%%% 30
\bibitem{Etazolate}
\begin{botherref}
A Study to Determine the Clinical Safety/Tolerability and Exploratory Efficacy
  of EHT 0202 as Adjunctive Therapy to Acetylcholinesterase Inhibitor in Mild
  to Moderate Alzheimer's Disease (EHT0202/002).
Accessed on 01.10.2022.
\url{https://clinicaltrials.gov/ct2/show/NCT00880412}
\end{botherref}
\endbibitem

%%% 31
\bibitem{fuji}
\begin{botherref}
\oauthor{\bsnm{Fujimaki}, \binits{T.}},
\oauthor{\bsnm{Saiki}, \binits{S.}},
\oauthor{\bsnm{Tashiro}, \binits{E.}},
\oauthor{\bsnm{Yamada}, \binits{D.}},
\oauthor{\bsnm{Kitagawa}, \binits{M.}},
\oauthor{\bsnm{Hattori}, \binits{N.}},
\oauthor{\bsnm{Imoto}, \binits{M.}}:
Identification of licopyranocoumarin and glycyrurol from herbal medicines as
  neuroprotective compounds for parkinson's disease.
PLoS One.
\textbf{9(6):e100395}
(2014)
\end{botherref}
\endbibitem

%%% 32
\bibitem{MEI2019107758}
\begin{barticle}
\bauthor{\bsnm{Mei}, \binits{M.}},
\bauthor{\bsnm{Zhou}, \binits{Y.}},
\bauthor{\bsnm{Liu}, \binits{M.}},
\bauthor{\bsnm{Zhao}, \binits{F.}},
\bauthor{\bsnm{Wang}, \binits{C.}},
\bauthor{\bsnm{Ding}, \binits{J.}},
\bauthor{\bsnm{Lu}, \binits{M.}},
\bauthor{\bsnm{Hu}, \binits{G.}}:
\batitle{Antioxidant and anti-inflammatory effects of dexrazoxane on
  dopaminergic neuron degeneration in rodent models of parkinson's disease}.
\bjtitle{Neuropharmacology}
\bvolume{160},
\bfpage{107758}
(\byear{2019}).
doi:\doiurl{10.1016/j.neuropharm.2019.107758}
\end{barticle}
\endbibitem

%%% 33
\bibitem{Brice2017}
\begin{botherref}
\oauthor{\bsnm{Owona}, \binits{B.}},
\oauthor{\bsnm{Zug}, \binits{C.}},
\oauthor{\bsnm{Schluesener}, \binits{H.}},
\oauthor{\bsnm{Zhang}, \binits{Z.}}:
Protective effects of forskolin on behavioral deficits and neuropathological
  changes in a mouse model of cerebral amyloidosis.
Journal of neuropathology and experimental neurology
(2016)
\end{botherref}
\endbibitem

%%% 34
\bibitem{do}
\begin{botherref}
\oauthor{\bsnm{Carreiras}, \binits{M.}},
\oauthor{\bsnm{Ismaili}, \binits{L.}},
\oauthor{\bsnm{Marco-Contelles}, \binits{J.}}:
Propargylamine-derived multi-target directed ligands for alzheimer's disease
  therapy.
Bioorg Med Chem Lett.
\textbf{30(3)}
(2020)
\end{botherref}
\endbibitem

%%% 35
\bibitem{Amit}
\begin{botherref}
\oauthor{\bsnm{Amit}, \binits{T.}},
\oauthor{\bsnm{Bar-Am}, \binits{O.}},
\oauthor{\bsnm{Mechlovich}, \binits{D.}},
\oauthor{\bsnm{Kupershmidt}, \binits{L.}},
\oauthor{\bsnm{Youdim}, \binits{M.}},
\oauthor{\bsnm{O.}, \binits{W.}}:
The novel multitarget iron chelating and propargylamine drug m30 affects app
  regulation and processing activities in alzheimer's disease models.
Neuropharmacology.
\textbf{123}
(2017)
\end{botherref}
\endbibitem

%%% 36
\bibitem{ou}
\begin{botherref}
\oauthor{\bsnm{Ou}, \binits{H.}},
\oauthor{\bsnm{Chien}, \binits{W.}},
\oauthor{\bsnm{Chung}, \binits{C.}},
\oauthor{\bsnm{Chang}, \binits{H.}},
\oauthor{\bsnm{Kao}, \binits{Y.}},
\oauthor{\bsnm{Wu}, \binits{P.}},
\oauthor{\bsnm{Tzeng}, \binits{N.}}:
Association between antibiotic treatment of chlamydia pneumoniae and reduced
  risk of alzheimer dementia: A nationwide cohort study in taiwan.
Frontiers in Aging Neuroscience
\textbf{13}
(2021).
doi:\doiurl{10.3389/fnagi.2021.701899}
\end{botherref}
\endbibitem

%%% 37
\bibitem{Angelucci}
\begin{botherref}
\oauthor{\bsnm{Angelucci}, \binits{F.}},
\oauthor{\bsnm{Cechova}, \binits{K.}},
\oauthor{\bsnm{Amlerova}, \binits{J.}},
\oauthor{\bsnm{Hort}, \binits{J.}}:
Antibiotics, gut microbiota, and alzheimer’s disease.
J Neuroinflammation.
\textbf{16108}
(2019)
\end{botherref}
\endbibitem

%%% 38
\bibitem{Jaturapatporn}
\begin{botherref}
\oauthor{\bsnm{Jaturapatporn}, \binits{D.}},
\oauthor{\bsnm{Isaac}, \binits{M.}},
\oauthor{\bsnm{McCleery}, \binits{J.}},
\oauthor{\bsnm{Tabet}, \binits{N.}}:
Aspirin, steroidal and non-steroidal anti-inflammatory drugs for the treatment
  of alzheimer's disease.
Cochrane Database Syst Rev.
\textbf{(2):CD006378}
(2012)
\end{botherref}
\endbibitem

%%% 39
\bibitem{Lara2003}
\begin{botherref}
\oauthor{\bsnm{Lara}, \binits{D.}},
\oauthor{\bsnm{Cruz}, \binits{M.R.}},
\oauthor{\bsnm{Xavier}, \binits{F.}},
\oauthor{\bsnm{Souza}, \binits{D.}},
\oauthor{\bsnm{Moriguchi}, \binits{E.}}:
Allopurinol for the treatment of aggressive behaviour in patients with
  dementia.
Int Clin Psychopharmacol.
(2003 Jan)
\end{botherref}
\endbibitem

%%% 40
\bibitem{Tikhonova2021}
\begin{botherref}
\oauthor{\bsnm{Tikhonova}, \binits{M.A.}},
\oauthor{\bsnm{Amstislavskaya}, \binits{T.G.}},
\oauthor{\bsnm{Ho}, \binits{Y.}},
\oauthor{\bsnm{Akopyan}, \binits{A.A.}},
\oauthor{\bsnm{Tenditnik}, \binits{M.V.}},
\oauthor{\bsnm{Ovsyukova}, \binits{M.V.}},
\oauthor{\bsnm{Bashirzade}, \binits{A.} \bsuffix{A}},
\oauthor{\bsnm{Dubrovina}, \binits{N.I.}},
\oauthor{\bsnm{Aftanas}, \binits{L.I.}}:
Neuroprotective effects of ceftriaxone involve the reduction of a$\beta$ burden
  and neuroinflammatory response in a mouse model of alzheimer’s disease.
Frontiers in Neuroscience
\textbf{15}
(2021).
doi:\doiurl{10.3389/fnins.2021.736786}
\end{botherref}
\endbibitem

%%% 41
\bibitem{Fernando}
\begin{botherref}
\oauthor{\bsnm{Fernando}, \binits{W.}},
\oauthor{\bsnm{R}, \binits{S.}},
\oauthor{\bsnm{Stephanie}, \binits{R.}},
\oauthor{\bsnm{Gardener}, \binits{S.}},
\oauthor{\bsnm{Villemagne}, \binits{V.}},
\oauthor{\bsnm{Burnham}, \binits{S.}},
\oauthor{\bsnm{Macaulay}, \binits{S.L.}},
\oauthor{\bsnm{Brown}, \binits{B.}},
\oauthor{\bsnm{Gupta}, \binits{V.B.}},
\oauthor{\bsnm{Sohrabi}, \binits{H.}},
\oauthor{\bsnm{Weinborn}, \binits{M.}},
\oauthor{\bsnm{Taddei}, \binits{K.}},
\oauthor{\bsnm{Laws}, \binits{S.}},
\oauthor{\bsnm{Goozee}, \binits{K.}},
\oauthor{\bsnm{Ames}, \binits{D.}},
\oauthor{\bsnm{Fowler}, \binits{C.}},
\oauthor{\bsnm{Maruff}, \binits{P.}},
\oauthor{\bsnm{Masters}, \binits{C.}},
\oauthor{\bsnm{Salvado}, \binits{O.}},
\oauthor{\bsnm{Rowe}, \binits{C.}},
\oauthor{\bsnm{Martins}, \binits{R.}}:
Associations of dietary protein and fiber intake with brain and blood
  amyloid-$\beta$.
J Alzheimers Dis.
\textbf{61(4)}
(2018)
\end{botherref}
\endbibitem

%%% 42
\bibitem{kakutani}
\begin{botherref}
\oauthor{\bsnm{Kakutani}, \binits{S.}},
\oauthor{\bsnm{Watanabe}, \binits{H.}},
\oauthor{\bsnm{Murayama}, \binits{N.}}:
Green tea intake and risks for dementia, alzheimer's disease, mild cognitive
  impairment, and cognitive impairment: A systematic review.
Nutrients
(2019)
\end{botherref}
\endbibitem

%%% 43
\bibitem{Baranowska}
\begin{botherref}
\oauthor{\bsnm{Baranowska-Wójcik}, \binits{E.}},
\oauthor{\bsnm{Szwajgier}, \binits{D.}},
\oauthor{\bsnm{Winiarska-Mieczan}, \binits{A.}}:
Honey as the potential natural source of cholinesterase inhibitors in
  alzheimer's disease.
\textbf{75(1)}
(2020)
\end{botherref}
\endbibitem

%%% 44
\bibitem{Maija}
\begin{botherref}
\oauthor{\bsnm{Ylilauri}, \binits{M.}},
\oauthor{\bsnm{Voutilainen}, \binits{S.}},
\oauthor{\bsnm{Eija}, \binits{L.}},
\oauthor{\bsnm{Virtanen}, \binits{H.E.K.}},
\oauthor{\bsnm{Tuomainen}, \binits{T.}},
\oauthor{\bsnm{Salonen}, \binits{J.}},
\oauthor{\bsnm{Virtanen}, \binits{J.}}:
Associations of dietary choline intake with risk of incident dementia and with
  cognitive performance: the kuopio ischaemic heart disease risk factor study.
The American journal of clinical nutrition
(2019)
\end{botherref}
\endbibitem

%%% 45
\bibitem{vintimilla2018}
\begin{botherref}
\oauthor{\bsnm{Vintimilla}, \binits{R.M.}},
\oauthor{\bsnm{Large}, \binits{S.E.}},
\oauthor{\bsnm{Gamboa}, \binits{A.}},
\oauthor{\bsnm{Rohlfing}, \binits{G.D.}},
\oauthor{\bsnm{O'Jile}, \binits{J.R.}},
\oauthor{\bsnm{Hall}, \binits{J.R.}},
\oauthor{\bsnm{O'Bryant}, \binits{S.E.}},
\oauthor{\bsnm{Johnson}, \binits{L.A.}}:
The link between potassium and mild cognitive impairment in mexican-americans.
Dement Geriatr Cogn Dis Extra.
(2018 Apr 24)
\end{botherref}
\endbibitem

%%% 46
\bibitem{Okuda}
\begin{botherref}
\oauthor{\bsnm{Okuda}, \binits{M.}},
\oauthor{\bsnm{Fujita}, \binits{Y.}},
\oauthor{\bsnm{Katsube}, \binits{T.}},
\oauthor{\bsnm{Tabata}, \binits{H.}},
\oauthor{\bsnm{Yoshino}, \binits{K.}},
\oauthor{\bsnm{Hashimoto}, \binits{M.}},
\oauthor{\bsnm{Sugimoto}, \binits{H.}}:
Highly water pressurized brown rice improves cognitive dysfunction in
  senescence-accelerated mouse prone 8 and reduces amyloid beta in the brain.
BMC Complement Altern Med.
\textbf{68(1):110}
(2018)
\end{botherref}
\endbibitem

%%% 47
\bibitem{Eskelinen2010}
\begin{botherref}
\oauthor{\bsnm{Eskelinen}, \binits{M.H.}},
\oauthor{\bsnm{Kivipelto}, \binits{M.}}:
Caffeine as a protective factor in dementia and alzheimer's disease.
J Alzheimers Dis
(2010)
\end{botherref}
\endbibitem

%%% 48
\bibitem{Smalheiser2019}
\begin{botherref}
\oauthor{\bsnm{Smalheiser}, \binits{N.R.}}:
Ketamine: A neglected therapy for alzheimer disease.
Frontiers in Aging Neuroscience
\textbf{11}
(2019).
doi:\doiurl{10.3389/fnagi.2019.00186}
\end{botherref}
\endbibitem

%%% 49
\bibitem{Hara2017}
\begin{botherref}
\oauthor{\bsnm{Hara}, \binits{Y.}},
\oauthor{\bsnm{McKeehan}, \binits{N.}},
\oauthor{\bsnm{Dacks}, \binits{P.A.}},
\oauthor{\bsnm{Fillit}, \binits{H.M.}}:
Evaluation of the neuroprotective potential of n-acetylcysteine for prevention
  and treatment of cognitive aging and dementia.
J Prev Alzheimers Dis.
(2017)
\end{botherref}
\endbibitem

%%% 50
\bibitem{Kilicoglu}
\begin{botherref}
\oauthor{\bsnm{Kilicoglu}, \binits{H.}},
\oauthor{\bsnm{Shin}, \binits{D.}},
\oauthor{\bsnm{Fiszman}, \binits{M.}},
\oauthor{\bsnm{Rosemblat}, \binits{G.}},
\oauthor{\bsnm{Rindflesch}, \binits{T.}}:
Semmeddb: a pubmed-scale repository of biomedical semantic predications.
Bioinformatics
\textbf{28(23)}
(2012)
\end{botherref}
\endbibitem

%%% 51
\bibitem{Kilicoglu1}
\begin{botherref}
\oauthor{\bsnm{Kilicoglu}, \binits{H.}},
\oauthor{\bsnm{Rosemblat}, \binits{G.}},
\oauthor{\bsnm{M}, \binits{F.}},
\oauthor{\bsnm{Shin}, \binits{D.}}:
Broad-coverage biomedical relation extraction with semrep.
BMC Bioinformatics
\textbf{21:1-28.}
(2020)
\end{botherref}
\endbibitem

%%% 52
\bibitem{McInnes}
\begin{botherref}
\oauthor{\bsnm{McInnes}, \binits{B.T.}}:
Extending the log-likelihood measure to improve collocation identification.
Master's thesis,
Univerity of Minnesota, Minneapolis
(2004)
\end{botherref}
\endbibitem

%%% 53
\bibitem{Lin}
\begin{bchapter}
\bauthor{\bsnm{Lin}, \binits{Y.}},
\bauthor{\bsnm{Liu}, \binits{Z.}},
\bauthor{\bsnm{Sun}, \binits{M.}},
\bauthor{\bsnm{Liu}, \binits{Y.}},
\bauthor{\bsnm{Zhu}, \binits{X.}}:
\bctitle{Learning entity and relation embeddings for knowledge graph
  completion}.
In: \bbtitle{Proceedings of the Twenty-Ninth AAAI Conference on Artificial
  Intelligence}.
\bsertitle{AAAI'15},
pp. \bfpage{2181}--\blpage{2187}.
\bpublisher{AAAI Press},
\blocation{Austin, Texas}
(\byear{2015})
\end{bchapter}
\endbibitem

%%% 54
\bibitem{DGL-KE}
\begin{bchapter}
\bauthor{\bsnm{Zheng}, \binits{D.}},
\bauthor{\bsnm{Song}, \binits{X.}},
\bauthor{\bsnm{Ma}, \binits{C.}},
\bauthor{\bsnm{Tan}, \binits{Z.}},
\bauthor{\bsnm{Ye}, \binits{Z.}},
\bauthor{\bsnm{Dong}, \binits{J.}},
\bauthor{\bsnm{Xiong}, \binits{H.}},
\bauthor{\bsnm{Zhang}, \binits{Z.}},
\bauthor{\bsnm{Karypis}, \binits{G.}}:
\bctitle{Dgl-ke: Training knowledge graph embeddings at scale}.
In: \bbtitle{Proceedings of the 43rd International ACM SIGIR Conference on
  Research and Development in Information Retrieval}.
\bsertitle{SIGIR '20},
pp. \bfpage{739}--\blpage{748}.
\bpublisher{Association for Computing Machinery},
\blocation{New York, NY, USA}
(\byear{2020})
\end{bchapter}
\endbibitem

%%% 55
\bibitem{Bordes}
\begin{botherref}
\oauthor{\bsnm{Bordes}, \binits{A.}},
\oauthor{\bsnm{Usunier}, \binits{N.}},
\oauthor{\bsnm{Garcia-Duran}, \binits{A.}},
\oauthor{\bsnm{Weston}, \binits{J.}},
\oauthor{\bsnm{Yakhnenko}, \binits{O.}}:
Translating embeddings for modeling multi-relational data
\textbf{26}
(2013)
\end{botherref}
\endbibitem

%%% 56
\bibitem{yang}
\begin{botherref}
\oauthor{\bsnm{Yang}, \binits{B.}},
\oauthor{\bsnm{Yih}, \binits{W.}},
\oauthor{\bsnm{X}, \binits{H.}},
\oauthor{\bsnm{Gao}, \binits{J.}},
\oauthor{\bsnm{Deng}, \binits{L.}}:
Embedding entities and relations for learning and inference in knowledge bases
(2015).
\arxivurl{1412.6575}
\end{botherref}
\endbibitem

%%% 57
\bibitem{trouillon2016complex}
\begin{botherref}
\oauthor{\bsnm{Trouillon}, \binits{T.}},
\oauthor{\bsnm{Welbl}, \binits{J.}},
\oauthor{\bsnm{Riedel}, \binits{S.}},
\oauthor{\bsnm{Gaussier}, \binits{E.}},
\oauthor{\bsnm{Bouchard}, \binits{G.}}:
Complex embeddings for simple link prediction
(2016).
\arxivurl{1606.06357}
\end{botherref}
\endbibitem

%%% 58
\bibitem{HENRY201720}
\begin{barticle}
\bauthor{\bsnm{Henry}, \binits{S.}},
\bauthor{\bsnm{McInnes}, \binits{B.T.}}:
\batitle{Literature based discovery: Models, methods, and trends}.
\bjtitle{Journal of Biomedical Informatics}
\bvolume{74},
\bfpage{20}--\blpage{32}
(\byear{2017}).
doi:\doiurl{10.1016/j.jbi.2017.08.011}
\end{barticle}
\endbibitem

\end{thebibliography}

\newcommand{\BMCxmlcomment}[1]{}

\BMCxmlcomment{

<refgrp>

<bibl id="B1">
  <title><p>Neurodegenerative diseases Latest research and news</p></title>
  <url>https://www.nature.com/subjects/neurodegenerative-diseases</url>
  <note>Accessed on 01.10.2022</note>
</bibl>

<bibl id="B2">
  <title><p>Neurodegeneration and Alzheimer's disease (AD). What Can Proteomics
  Tell Us About the Alzheimer's Brain?.</p></title>
  <aug>
    <au><snm>Moya Alvarado</snm><fnm>G</fnm></au>
    <au><snm>Gershoni Emek</snm><fnm>N</fnm></au>
    <au><snm>Perlson</snm><fnm>E</fnm></au>
    <au><snm>Bronfman</snm><fnm>FC</fnm></au>
  </aug>
  <source>Mol Cell Proteomics.</source>
  <pubdate>2016</pubdate>
  <volume>15(2)</volume>
</bibl>

<bibl id="B3">
  <title><p>Learning from electronic health records across multiple sites: A
  communication-efficient and privacy-preserving distributed
  algorithm.</p></title>
  <aug>
    <au><snm>Duan</snm><fnm>R</fnm></au>
    <au><snm>Boland</snm><fnm>MR</fnm></au>
    <au><snm>Liu</snm><fnm>Z</fnm></au>
    <au><snm>Liu</snm><fnm>Y</fnm></au>
    <au><snm>Chang</snm><fnm>HH</fnm></au>
    <au><snm>Xu</snm><fnm>H</fnm></au>
    <au><snm>Chu</snm><fnm>H</fnm></au>
    <au><snm>Schmid</snm><fnm>CH</fnm></au>
    <au><snm>Forrest</snm><fnm>CH</fnm></au>
    <au><snm>Holmes</snm><fnm>JH</fnm></au>
    <au><snm>Schuemie</snm><fnm>MJ</fnm></au>
    <au><snm>Berlin</snm><fnm>JA</fnm></au>
    <au><snm>Moore</snm><fnm>JH</fnm></au>
    <au><snm>Chen</snm><fnm>Y</fnm></au>
  </aug>
  <source>J Am Med Inform Assoc.</source>
  <pubdate>2020</pubdate>
  <volume>27(3)</volume>
</bibl>

<bibl id="B4">
  <title><p>Drug repositioning: identifying and developing new uses for
  existing drugs.</p></title>
  <aug>
    <au><snm>Ashburn</snm><fnm>T</fnm></au>
    <au><snm>Thor</snm><fnm>K</fnm></au>
  </aug>
  <source>Nat Rev Drug Discov</source>
  <pubdate>2004</pubdate>
  <volume>3</volume>
</bibl>

<bibl id="B5">
  <title><p>A review of computational drug repurposing</p></title>
  <aug>
    <au><snm>Kyungsoo</snm><fnm>P</fnm></au>
  </aug>
  <source>tcp</source>
  <pubdate>2019</pubdate>
  <volume>27</volume>
  <issue>2</issue>
  <fpage>59</fpage>
  <lpage>63</lpage>
  <url>http://www.e-sciencecentral.org/articles/?scid=1130297</url>
</bibl>

<bibl id="B6">
  <title><p>Drug repurposing: progress, challenges and
  recommendations.</p></title>
  <aug>
    <au><snm>Pushpakom</snm><fnm>S</fnm></au>
    <au><snm>Iorio</snm><fnm>F</fnm></au>
    <au><snm>Eyers</snm><fnm>PA</fnm></au>
    <au><snm>Escott</snm><fnm>KJ</fnm></au>
    <au><snm>Hopper</snm><fnm>S</fnm></au>
    <au><snm>Wells</snm><fnm>A</fnm></au>
    <au><snm>Doig</snm><fnm>A</fnm></au>
    <au><snm>Guilliams</snm><fnm>T</fnm></au>
    <au><snm>Latimer</snm><fnm>J</fnm></au>
    <au><snm>McNamee</snm><fnm>C</fnm></au>
    <au><snm>Norris</snm><fnm>A</fnm></au>
    <au><snm>Sanseau</snm><fnm>P</fnm></au>
    <au><snm>Cavalla</snm><fnm>D</fnm></au>
    <au><snm>Pirmohamed</snm><fnm>M</fnm></au>
  </aug>
  <source>Nat Rev Drug Discov.</source>
  <pubdate>2019</pubdate>
  <volume>18(1)</volume>
</bibl>

<bibl id="B7">
  <title><p>A review of biomedical datasets relating to drug discovery: A
  knowledge graph perspective</p></title>
  <aug>
    <au><snm>Bonner</snm><fnm>S</fnm></au>
    <au><snm>Barrett</snm><fnm>IP</fnm></au>
    <au><snm>Ye</snm><fnm>C</fnm></au>
    <au><snm>Swiers</snm><fnm>R</fnm></au>
    <au><snm>Engkvist</snm><fnm>O</fnm></au>
    <au><snm>Bender</snm><fnm>A</fnm></au>
    <au><snm>Hoyt</snm><fnm>CT</fnm></au>
    <au><snm>Hamilton</snm><fnm>W</fnm></au>
  </aug>
  <source>arXiv preprint arXiv:2102.10062</source>
  <pubdate>2021</pubdate>
</bibl>

<bibl id="B8">
  <title><p>Drug repurposing for COVID-19 via knowledge graph
  completion</p></title>
  <aug>
    <au><snm>Zhang</snm><fnm>R</fnm></au>
    <au><snm>Hristovski</snm><fnm>D</fnm></au>
    <au><snm>Schutte</snm><fnm>D</fnm></au>
    <au><snm>Kastrin</snm><fnm>A</fnm></au>
    <au><snm>Fiszman</snm><fnm>M</fnm></au>
    <au><snm>Kilicoglu</snm><fnm>H</fnm></au>
  </aug>
  <source>Journal of biomedical informatics</source>
  <pubdate>2021</pubdate>
  <volume>115, p.103696.</volume>
</bibl>

<bibl id="B9">
  <title><p>Drug Repurposing for the Treatment of COVID-19: A Knowledge Graph
  Approach.</p></title>
  <aug>
    <au><snm>Yan</snm><fnm>VKC</fnm></au>
    <au><snm>Li</snm><fnm>X</fnm></au>
    <au><snm>Ye</snm><fnm>X</fnm></au>
    <au><snm>Ou</snm><fnm>M</fnm></au>
    <au><snm>Luo</snm><fnm>R</fnm></au>
    <au><snm>Zhang</snm><fnm>Q</fnm></au>
    <au><snm>Tang</snm><fnm>B</fnm></au>
    <au><snm>BJ</snm><fnm>C</fnm></au>
    <au><snm>I</snm><fnm>H</fnm></au>
    <au><snm>Siu</snm><fnm>CW</fnm></au>
    <au><snm>ICK</snm><fnm>W</fnm></au>
    <au><snm>RCK</snm><fnm>C</fnm></au>
    <au><snm>EW</snm><fnm>C</fnm></au>
  </aug>
  <source>Epub ahead of print. Erratum in: Adv Ther (Weinh).</source>
  <pubdate>2021</pubdate>
  <volume>4(10):2100179</volume>
</bibl>

<bibl id="B10">
  <title><p>Knowledge Graph-Based Approaches to Drug Repurposing for
  COVID-19</p></title>
  <aug>
    <au><snm>Al Saleem</snm><fnm>J</fnm></au>
    <au><snm>Granet</snm><fnm>R</fnm></au>
    <au><snm>Ramakrishnan</snm><fnm>S</fnm></au>
    <au><snm>Ciancetta</snm><fnm>NA</fnm></au>
    <au><snm>Saveson</snm><fnm>C</fnm></au>
    <au><snm>Gessner</snm><fnm>C</fnm></au>
    <au><snm>Zhou</snm><fnm>Q</fnm></au>
  </aug>
  <source>Journal of Chemical Information and Modeling</source>
  <pubdate>2021</pubdate>
  <volume>61</volume>
  <issue>8</issue>
  <fpage>4058</fpage>
  <lpage>4067</lpage>
  <url>https://doi.org/10.1021/acs.jcim.1c00642</url>
  <note>PMID: 34297570</note>
</bibl>

<bibl id="B11">
  <title><p>A Literature-Based Knowledge Graph Embedding Method for Identifying
  Drug Repurposing Opportunities in Rare Diseases.</p></title>
  <aug>
    <au><snm>Sosa</snm><fnm>DN</fnm></au>
    <au><snm>Derry</snm><fnm>A</fnm></au>
    <au><snm>Guo</snm><fnm>M</fnm></au>
    <au><snm>Wei</snm><fnm>E</fnm></au>
    <au><snm>Brinton</snm><fnm>C</fnm></au>
    <au><snm>Altman</snm><fnm>RB</fnm></au>
  </aug>
  <source>Pac Symp Biocomput</source>
  <pubdate>2020</pubdate>
  <volume>25</volume>
</bibl>

<bibl id="B12">
  <title><p>Drug prioritization using the semantic properties of a knowledge
  graph.</p></title>
  <aug>
    <au><snm>Malas</snm><fnm>TB</fnm></au>
    <au><snm>Vlietstra</snm><fnm>WJ</fnm></au>
    <au><snm>Kudrin</snm><fnm>R</fnm></au>
    <au><snm>Starikov</snm><fnm>S</fnm></au>
    <au><snm>Charrout</snm><fnm>M</fnm></au>
    <au><snm>Roos</snm><fnm>M</fnm></au>
    <au><snm>Peters</snm><fnm>DJM</fnm></au>
    <au><snm>Kors</snm><fnm>JA</fnm></au>
    <au><snm>Vos</snm><fnm>R</fnm></au>
    <au><snm>PAC</snm><fnm>H</fnm></au>
    <au><snm>Mulligen</snm><fnm>EM</fnm></au>
    <au><snm>Hettne</snm><fnm>KM</fnm></au>
  </aug>
  <source>Sci Rep.</source>
  <pubdate>2019</pubdate>
  <volume>9(1):6281</volume>
</bibl>

<bibl id="B13">
  <title><p>Nutrition, brain aging, and neurodegeneration</p></title>
  <aug>
    <au><snm>Joseph</snm><fnm>J</fnm></au>
    <au><snm>Cole</snm><fnm>G</fnm></au>
    <au><snm>Head</snm><fnm>E</fnm></au>
    <au><snm>D</snm><fnm>I</fnm></au>
  </aug>
  <source>Journal of Neuroscience</source>
  <pubdate>2009</pubdate>
  <volume>29</volume>
</bibl>

<bibl id="B14">
  <title><p>PubMed</p></title>
  <url>https://pubmed.ncbi.nlm.nih.gov/</url>
  <note>Accessed on 01.10.2022</note>
</bibl>

<bibl id="B15">
  <title><p>{iDISK: the integrated DIetary Supplements Knowledge
  base}</p></title>
  <aug>
    <au><snm>Rizvi</snm><fnm>RF</fnm></au>
    <au><snm>Vasilakes</snm><fnm>J</fnm></au>
    <au><snm>Adam</snm><fnm>TJ</fnm></au>
    <au><snm>Melton</snm><fnm>GB</fnm></au>
    <au><snm>Bishop</snm><fnm>JR</fnm></au>
    <au><snm>Bian</snm><fnm>J</fnm></au>
    <au><snm>Tao</snm><fnm>C</fnm></au>
    <au><snm>Zhang</snm><fnm>R</fnm></au>
  </aug>
  <source>Journal of the American Medical Informatics Association</source>
  <pubdate>2020</pubdate>
  <volume>27</volume>
  <issue>4</issue>
  <fpage>539</fpage>
  <lpage>548</lpage>
</bibl>

<bibl id="B16">
  <title><p>Intrathecal corticosteroids might slow Alzheimer's disease
  progression.</p></title>
  <aug>
    <au><snm>Alisky</snm><fnm>JM</fnm></au>
  </aug>
  <source>Neuropsychiatric disease and treatment.</source>
  <pubdate>2008</pubdate>
  <volume>4,5</volume>
</bibl>

<bibl id="B17">
  <title><p>Antiviral therapy: Valacyclovir Treatment of Alzheimer's Disease
  (VALAD) Trial: protocol for a randomised, double-blind,placebo-controlled,
  treatment trial</p></title>
  <aug>
    <au><snm>Devanand</snm><fnm>DP</fnm></au>
    <au><snm>Andrews</snm><fnm>H</fnm></au>
    <au><snm>Kreisl</snm><fnm>WC</fnm></au>
    <au><snm>Razlighi</snm><fnm>Q</fnm></au>
    <au><snm>Gershon</snm><fnm>A</fnm></au>
    <au><snm>Stern</snm><fnm>Y</fnm></au>
    <au><snm>Mintz</snm><fnm>A</fnm></au>
    <au><snm>Wisniewski</snm><fnm>T</fnm></au>
    <au><snm>Acosta</snm><fnm>E</fnm></au>
    <au><snm>Pollina</snm><fnm>J</fnm></au>
    <au><snm>Katsikoumbas</snm><fnm>M</fnm></au>
    <au><snm>Bell</snm><fnm>KL</fnm></au>
    <au><snm>Pelton</snm><fnm>GH</fnm></au>
    <au><snm>Deliyannides</snm><fnm>D</fnm></au>
    <au><snm>Prasad</snm><fnm>KM</fnm></au>
    <au><snm>Huey</snm><fnm>ED</fnm></au>
  </aug>
  <source>BMJ Open</source>
  <pubdate>2020</pubdate>
  <volume>10(2):e0321112</volume>
</bibl>

<bibl id="B18">
  <title><p>Synergistic depletion of gut microbial consortia, but not
  individual antibiotics, reduces amyloidosis in APPPS1-21 Alzheimer's
  transgenic mice.</p></title>
  <aug>
    <au><snm>Dodiya</snm><fnm>HB</fnm></au>
    <au><snm>Frith</snm><fnm>M</fnm></au>
    <au><snm>Sidebottom</snm><fnm>A</fnm></au>
    <au><snm>Cao</snm><fnm>Y</fnm></au>
    <au><snm>Koval</snm><fnm>J</fnm></au>
    <au><snm>Chang</snm><fnm>E</fnm></au>
    <au><snm>Sisodia</snm><fnm>SS</fnm></au>
  </aug>
  <source>Scientific reports.</source>
  <pubdate>2020</pubdate>
</bibl>

<bibl id="B19">
  <title><p>A Pilot Open Labeled Study of Tacrolimus in Alzheimer's
  Disease</p></title>
  <url>https://clinicaltrials.gov/ct2/show/results/NCT04263519</url>
  <note>Accessed on 01.10.2022</note>
</bibl>

<bibl id="B20">
  <title><p>Intrathecal corticosteroids might slow Alzheimer's disease
  progression.</p></title>
  <aug>
    <au><snm>Alisky</snm><fnm>JM</fnm></au>
  </aug>
  <source>Neuropsychiatr Dis Treat</source>
  <pubdate>2008</pubdate>
  <volume>4(5)</volume>
</bibl>

<bibl id="B21">
  <title><p>The Amyloid Cascade Hypothesis in Alzheimer's Disease: It's Time to
  Change Our Mind</p></title>
  <aug>
    <au><snm>Ricciarelli</snm><fnm>R</fnm></au>
    <au><snm>Fedele</snm><fnm>E</fnm></au>
  </aug>
  <source>Curr Neuropharmacol</source>
  <pubdate>2017</pubdate>
  <volume>15(6)</volume>
</bibl>

<bibl id="B22">
  <title><p>Propranolol reduces cognitive deficits, amyloid and tau pathology
  in Alzheimer's transgenic mice.</p></title>
  <aug>
    <au><snm>Dobarro</snm><fnm>M</fnm></au>
    <au><snm>Gerenu</snm><fnm>G</fnm></au>
    <au><snm>Ramírez</snm><fnm>MJ</fnm></au>
  </aug>
  <source>Int J Neuropsychopharmacol</source>
  <pubdate>2013</pubdate>
  <volume>16(10)</volume>
</bibl>

<bibl id="B23">
  <title><p>Amifostine (WR-2721) Mitigates Cognitive Injury Induced by Heavy
  Ion Radiation in Male Mice and Alters Behavior and Brain
  Connectivity</p></title>
  <aug>
    <au><snm>Boutros</snm><fnm>SW</fnm></au>
    <au><snm>Zimmerman</snm><fnm>B</fnm></au>
    <au><snm>Nagy</snm><fnm>SC</fnm></au>
    <au><snm>Lee</snm><fnm>JS</fnm></au>
    <au><snm>Perez</snm><fnm>R</fnm></au>
    <au><snm>Raber</snm><fnm>J</fnm></au>
  </aug>
  <source>Frontiers in Physiology</source>
  <pubdate>2021</pubdate>
  <volume>12</volume>
  <url>https://www.frontiersin.org/article/10.3389/fphys.2021.770502</url>
</bibl>

<bibl id="B24">
  <title><p>Betaine attenuates Alzheimer-like pathological changes and memory
  deficits induced by homocysteine.</p></title>
  <aug>
    <au><snm>Chai</snm><fnm>GS</fnm></au>
    <au><snm>Jiang</snm><fnm>X</fnm></au>
    <au><snm>Ni</snm><fnm>ZF</fnm></au>
    <au><snm>Ma</snm><fnm>ZW</fnm></au>
    <au><snm>Xie</snm><fnm>AJ</fnm></au>
    <au><snm>Cheng</snm><fnm>XS</fnm></au>
    <au><snm>Wang</snm><fnm>Q</fnm></au>
    <au><snm>Wang</snm><fnm>JZ</fnm></au>
    <au><snm>Liu</snm><fnm>GP</fnm></au>
  </aug>
  <source>J Neurochem.</source>
  <pubdate>2013, Feb</pubdate>
  <volume>(3):388-96.</volume>
</bibl>

<bibl id="B25">
  <title><p>Oxytocin in Alzheimer's disease: postmortem brain
  levels</p></title>
  <aug>
    <au><snm>Mazurek</snm><fnm>MF</fnm></au>
    <au><snm>Beal</snm><fnm>MF</fnm></au>
    <au><snm>Bird</snm><fnm>ED</fnm></au>
    <au><snm>Martin</snm><fnm>JB</fnm></au>
  </aug>
  <source>Neurology</source>
  <pubdate>1987</pubdate>
  <volume>37(6)</volume>
</bibl>

<bibl id="B26">
  <title><p>Oral microbiota and Alzheimer's disease: Do all roads lead to
  Rome?.</p></title>
  <aug>
    <au><snm>Sureda</snm><fnm>A</fnm></au>
    <au><snm>Daglia</snm><fnm>M</fnm></au>
    <au><snm>Castilla</snm><fnm>SA</fnm></au>
    <au><snm>Sanadgol</snm><fnm>N</fnm></au>
    <au><snm>Nabavi</snm><fnm>SF</fnm></au>
    <au><snm>Khan</snm><fnm>H</fnm></au>
    <au><snm>Belwal</snm><fnm>T</fnm></au>
    <au><snm>Jeandet</snm><fnm>P</fnm></au>
    <au><snm>Marchese</snm><fnm>A</fnm></au>
    <au><snm>Pistollato</snm><fnm>F</fnm></au>
    <au><snm>Forbes Hernandez</snm><fnm>T</fnm></au>
    <au><snm>Battino</snm><fnm>M</fnm></au>
    <au><snm>Berindan Neagoe</snm><fnm>I</fnm></au>
    <au><snm>G</snm><fnm>D</fnm></au>
    <au><snm>Nabavi</snm><fnm>SM</fnm></au>
  </aug>
  <source>Pharmacol Res</source>
  <pubdate>2020</pubdate>
  <volume>151:104582</volume>
</bibl>

<bibl id="B27">
  <title><p>Impact of minocycline on neurodegenerative diseases in rodents: a
  meta-analysis.</p></title>
  <aug>
    <au><snm>Li</snm><fnm>C</fnm></au>
    <au><snm>Yuan</snm><fnm>K</fnm></au>
    <au><snm>Schluesener</snm><fnm>H.</fnm></au>
  </aug>
  <source>Rev Neurosci.</source>
  <pubdate>2013</pubdate>
  <volume>24(5)</volume>
</bibl>

<bibl id="B28">
  <title><p>Tetracycline repurposing in neurodegeneration: focus on Parkinson's
  disease.</p></title>
  <aug>
    <au><snm>Bortolanza</snm><fnm>M</fnm></au>
    <au><snm>Nascimento</snm><fnm>GC</fnm></au>
    <au><snm>Socias</snm><fnm>SB</fnm></au>
    <au><snm>Ploper</snm><fnm>D</fnm></au>
    <au><snm>Chehín</snm><fnm>RN</fnm></au>
    <au><snm>Raisman Vozari</snm><fnm>R</fnm></au>
    <au><snm>Del Bel</snm><fnm>E</fnm></au>
  </aug>
  <source>J Neural Transm (Vienna).</source>
  <pubdate>2018</pubdate>
  <volume>125(10)</volume>
</bibl>

<bibl id="B29">
  <title><p>Implications of sodium hydrogen exchangers in various brain
  diseases.</p></title>
  <aug>
    <au><snm>Verma</snm><fnm>V</fnm></au>
    <au><snm>Bali</snm><fnm>A</fnm></au>
    <au><snm>Singh</snm><fnm>N</fnm></au>
    <au><snm>Jaggi</snm><fnm>AS</fnm></au>
  </aug>
  <source>J Basic Clin Physiol Pharmacol.</source>
  <pubdate>2015</pubdate>
  <volume>26(5)</volume>
</bibl>

<bibl id="B30">
  <title><p>A Study to Determine the Clinical Safety/Tolerability and
  Exploratory Efficacy of EHT 0202 as Adjunctive Therapy to
  Acetylcholinesterase Inhibitor in Mild to Moderate Alzheimer's Disease
  (EHT0202/002)</p></title>
  <url>https://clinicaltrials.gov/ct2/show/NCT00880412</url>
  <note>Accessed on 01.10.2022</note>
</bibl>

<bibl id="B31">
  <title><p>Identification of licopyranocoumarin and glycyrurol from herbal
  medicines as neuroprotective compounds for Parkinson's disease.</p></title>
  <aug>
    <au><snm>Fujimaki</snm><fnm>T</fnm></au>
    <au><snm>Saiki</snm><fnm>S</fnm></au>
    <au><snm>Tashiro</snm><fnm>E</fnm></au>
    <au><snm>Yamada</snm><fnm>D</fnm></au>
    <au><snm>Kitagawa</snm><fnm>M</fnm></au>
    <au><snm>Hattori</snm><fnm>N</fnm></au>
    <au><snm>Imoto</snm><fnm>M</fnm></au>
  </aug>
  <source>PLoS One.</source>
  <pubdate>2014</pubdate>
  <volume>9(6):e100395</volume>
</bibl>

<bibl id="B32">
  <title><p>Antioxidant and anti-inflammatory effects of dexrazoxane on
  dopaminergic neuron degeneration in rodent models of Parkinson's
  disease</p></title>
  <aug>
    <au><snm>Mei</snm><fnm>M</fnm></au>
    <au><snm>Zhou</snm><fnm>Y</fnm></au>
    <au><snm>Liu</snm><fnm>M</fnm></au>
    <au><snm>Zhao</snm><fnm>F</fnm></au>
    <au><snm>Wang</snm><fnm>C</fnm></au>
    <au><snm>Ding</snm><fnm>J</fnm></au>
    <au><snm>Lu</snm><fnm>M</fnm></au>
    <au><snm>Hu</snm><fnm>G</fnm></au>
  </aug>
  <source>Neuropharmacology</source>
  <pubdate>2019</pubdate>
  <volume>160</volume>
  <fpage>107758</fpage>
  <url>https://www.sciencedirect.com/science/article/pii/S002839081930317X</url>
</bibl>

<bibl id="B33">
  <title><p>Protective Effects of Forskolin on Behavioral Deficits and
  Neuropathological Changes in a Mouse Model of Cerebral
  Amyloidosis.</p></title>
  <aug>
    <au><snm>Owona</snm><fnm>BA</fnm></au>
    <au><snm>Zug</snm><fnm>C</fnm></au>
    <au><snm>Schluesener</snm><fnm>HJ</fnm></au>
    <au><snm>Zhang</snm><fnm>Z</fnm></au>
  </aug>
  <source>Journal of neuropathology and experimental neurology</source>
  <pubdate>2016</pubdate>
</bibl>

<bibl id="B34">
  <title><p>Propargylamine-derived multi-target directed ligands for
  Alzheimer's disease therapy</p></title>
  <aug>
    <au><snm>Carreiras</snm><fnm>MC</fnm></au>
    <au><snm>Ismaili</snm><fnm>L</fnm></au>
    <au><snm>Marco Contelles</snm><fnm>J</fnm></au>
  </aug>
  <source>Bioorg Med Chem Lett.</source>
  <pubdate>2020</pubdate>
  <volume>30(3)</volume>
</bibl>

<bibl id="B35">
  <title><p>The novel multitarget iron chelating and propargylamine drug M30
  affects APP regulation and processing activities in Alzheimer's disease
  models.</p></title>
  <aug>
    <au><snm>Amit</snm><fnm>T</fnm></au>
    <au><snm>Bar Am</snm><fnm>O</fnm></au>
    <au><snm>Mechlovich</snm><fnm>D</fnm></au>
    <au><snm>Kupershmidt</snm><fnm>L</fnm></au>
    <au><snm>Youdim</snm><fnm>MBH</fnm></au>
    <au><snm>O.</snm><fnm>W</fnm></au>
  </aug>
  <source>Neuropharmacology.</source>
  <pubdate>2017</pubdate>
  <volume>123</volume>
</bibl>

<bibl id="B36">
  <title><p>Association Between Antibiotic Treatment of Chlamydia pneumoniae
  and Reduced Risk of Alzheimer Dementia: A Nationwide Cohort Study in
  Taiwan</p></title>
  <aug>
    <au><snm>Ou</snm><fnm>H</fnm></au>
    <au><snm>Chien</snm><fnm>WC</fnm></au>
    <au><snm>Chung</snm><fnm>C</fnm></au>
    <au><snm>Chang</snm><fnm>H</fnm></au>
    <au><snm>Kao</snm><fnm>Y</fnm></au>
    <au><snm>Wu</snm><fnm>PC</fnm></au>
    <au><snm>Tzeng</snm><fnm>NS</fnm></au>
  </aug>
  <source>Frontiers in Aging Neuroscience</source>
  <pubdate>2021</pubdate>
  <volume>13</volume>
  <url>https://www.frontiersin.org/article/10.3389/fnagi.2021.701899</url>
</bibl>

<bibl id="B37">
  <title><p>Antibiotics, gut microbiota, and Alzheimer’s disease.</p></title>
  <aug>
    <au><snm>Angelucci</snm><fnm>F</fnm></au>
    <au><snm>Cechova</snm><fnm>K</fnm></au>
    <au><snm>Amlerova</snm><fnm>J</fnm></au>
    <au><snm>Hort</snm><fnm>J</fnm></au>
  </aug>
  <source>J Neuroinflammation.</source>
  <pubdate>2019</pubdate>
  <volume>16,108</volume>
</bibl>

<bibl id="B38">
  <title><p>Aspirin, steroidal and non-steroidal anti-inflammatory drugs for
  the treatment of Alzheimer's disease.</p></title>
  <aug>
    <au><snm>Jaturapatporn</snm><fnm>D</fnm></au>
    <au><snm>Isaac</snm><fnm>MG</fnm></au>
    <au><snm>McCleery</snm><fnm>J</fnm></au>
    <au><snm>Tabet</snm><fnm>N.</fnm></au>
  </aug>
  <source>Cochrane Database Syst Rev.</source>
  <pubdate>2012</pubdate>
  <volume>(2):CD006378</volume>
</bibl>

<bibl id="B39">
  <title><p>Allopurinol for the treatment of aggressive behaviour in patients
  with dementia.</p></title>
  <aug>
    <au><snm>Lara</snm><fnm>DR</fnm></au>
    <au><snm>Cruz</snm><fnm>M RS</fnm></au>
    <au><snm>Xavier</snm><fnm>F</fnm></au>
    <au><snm>Souza</snm><fnm>DO</fnm></au>
    <au><snm>Moriguchi</snm><fnm>EH</fnm></au>
  </aug>
  <source>Int Clin Psychopharmacol.</source>
  <pubdate>2003 Jan</pubdate>
</bibl>

<bibl id="B40">
  <title><p>Neuroprotective Effects of Ceftriaxone Involve the Reduction of
  A$\beta$ Burden and Neuroinflammatory Response in a Mouse Model of
  Alzheimer’s Disease</p></title>
  <aug>
    <au><snm>Tikhonova</snm><fnm>M A</fnm></au>
    <au><snm>Amstislavskaya</snm><fnm>T G.</fnm></au>
    <au><snm>Ho</snm><fnm>YJ</fnm></au>
    <au><snm>Akopyan</snm><fnm>A A.</fnm></au>
    <au><snm>Tenditnik</snm><fnm>M V.</fnm></au>
    <au><snm>Ovsyukova</snm><fnm>M V.</fnm></au>
    <au><snm>Bashirzade</snm><fnm>A.</fnm></au>
    <au><snm>Dubrovina</snm><fnm>N I.</fnm></au>
    <au><snm>Aftanas</snm><fnm>L I.</fnm></au>
  </aug>
  <source>Frontiers in Neuroscience</source>
  <pubdate>2021</pubdate>
  <volume>15</volume>
  <url>https://www.frontiersin.org/article/10.3389/fnins.2021.736786</url>
</bibl>

<bibl id="B41">
  <title><p>Associations of Dietary Protein and Fiber Intake with Brain and
  Blood Amyloid-$\beta$.</p></title>
  <aug>
    <au><snm>Fernando</snm><fnm>WMADB</fnm></au>
    <au><snm>R</snm><fnm>S</fnm></au>
    <au><snm>Stephanie</snm><fnm>R</fnm></au>
    <au><snm>Gardener</snm><fnm>S</fnm></au>
    <au><snm>Villemagne</snm><fnm>V</fnm></au>
    <au><snm>Burnham</snm><fnm>S</fnm></au>
    <au><snm>Macaulay</snm><fnm>S. L</fnm></au>
    <au><snm>Brown</snm><fnm>B</fnm></au>
    <au><snm>Gupta</snm><fnm>V B</fnm></au>
    <au><snm>Sohrabi</snm><fnm>H</fnm></au>
    <au><snm>Weinborn</snm><fnm>M</fnm></au>
    <au><snm>Taddei</snm><fnm>K</fnm></au>
    <au><snm>Laws</snm><fnm>S</fnm></au>
    <au><snm>Goozee</snm><fnm>K</fnm></au>
    <au><snm>Ames</snm><fnm>D</fnm></au>
    <au><snm>Fowler</snm><fnm>C</fnm></au>
    <au><snm>Maruff</snm><fnm>P</fnm></au>
    <au><snm>Masters</snm><fnm>C</fnm></au>
    <au><snm>Salvado</snm><fnm>O</fnm></au>
    <au><snm>Rowe</snm><fnm>C</fnm></au>
    <au><snm>Martins</snm><fnm>R</fnm></au>
  </aug>
  <source>J Alzheimers Dis.</source>
  <pubdate>2018</pubdate>
  <volume>61(4)</volume>
</bibl>

<bibl id="B42">
  <title><p>Green Tea Intake and Risks for Dementia, Alzheimer's Disease, Mild
  Cognitive Impairment, and Cognitive Impairment: A Systematic
  Review.</p></title>
  <aug>
    <au><snm>Kakutani</snm><fnm>S</fnm></au>
    <au><snm>Watanabe</snm><fnm>H</fnm></au>
    <au><snm>Murayama</snm><fnm>N</fnm></au>
  </aug>
  <source>Nutrients</source>
  <pubdate>2019</pubdate>
</bibl>

<bibl id="B43">
  <title><p>Honey as the Potential Natural Source of Cholinesterase Inhibitors
  in Alzheimer's Disease.</p></title>
  <aug>
    <au><snm>Baranowska Wójcik</snm><fnm>E</fnm></au>
    <au><snm>Szwajgier</snm><fnm>D</fnm></au>
    <au><snm>Winiarska Mieczan</snm><fnm>A</fnm></au>
  </aug>
  <publisher>Plant Foods Hum Nutr.</publisher>
  <pubdate>2020</pubdate>
  <volume>75(1)</volume>
</bibl>

<bibl id="B44">
  <title><p>Associations of dietary choline intake with risk of incident
  dementia and with cognitive performance: the Kuopio Ischaemic Heart Disease
  Risk Factor Study.</p></title>
  <aug>
    <au><snm>Ylilauri</snm><fnm>MPT</fnm></au>
    <au><snm>Voutilainen</snm><fnm>S</fnm></au>
    <au><snm>Eija</snm><fnm>L</fnm></au>
    <au><snm>Virtanen</snm><fnm>HEK</fnm></au>
    <au><snm>Tuomainen</snm><fnm>TP</fnm></au>
    <au><snm>Salonen</snm><fnm>JT</fnm></au>
    <au><snm>Virtanen</snm><fnm>JK</fnm></au>
  </aug>
  <source>The American journal of clinical nutrition</source>
  <pubdate>2019</pubdate>
</bibl>

<bibl id="B45">
  <title><p>The Link between Potassium and Mild Cognitive Impairment in
  Mexican-Americans.</p></title>
  <aug>
    <au><snm>Vintimilla</snm><fnm>R M</fnm></au>
    <au><snm>Large</snm><fnm>S E</fnm></au>
    <au><snm>Gamboa</snm><fnm>A</fnm></au>
    <au><snm>Rohlfing</snm><fnm>G D</fnm></au>
    <au><snm>O'Jile</snm><fnm>J R</fnm></au>
    <au><snm>Hall</snm><fnm>J R</fnm></au>
    <au><snm>O'Bryant</snm><fnm>S E</fnm></au>
    <au><snm>Johnson</snm><fnm>L A</fnm></au>
  </aug>
  <source>Dement Geriatr Cogn Dis Extra.</source>
  <pubdate>2018 Apr 24</pubdate>
</bibl>

<bibl id="B46">
  <title><p>Highly water pressurized brown rice improves cognitive dysfunction
  in senescence-accelerated mouse prone 8 and reduces amyloid beta in the
  brain.</p></title>
  <aug>
    <au><snm>Okuda</snm><fnm>M</fnm></au>
    <au><snm>Fujita</snm><fnm>Y</fnm></au>
    <au><snm>Katsube</snm><fnm>T</fnm></au>
    <au><snm>Tabata</snm><fnm>H</fnm></au>
    <au><snm>Yoshino</snm><fnm>K</fnm></au>
    <au><snm>Hashimoto</snm><fnm>M</fnm></au>
    <au><snm>Sugimoto</snm><fnm>H</fnm></au>
  </aug>
  <source>BMC Complement Altern Med.</source>
  <pubdate>2018</pubdate>
  <volume>68(1):110</volume>
</bibl>

<bibl id="B47">
  <title><p>Caffeine as a protective factor in dementia and Alzheimer's
  disease.</p></title>
  <aug>
    <au><snm>Eskelinen</snm><fnm>M H</fnm></au>
    <au><snm>Kivipelto</snm><fnm>M</fnm></au>
  </aug>
  <source>J Alzheimers Dis</source>
  <pubdate>2010</pubdate>
</bibl>

<bibl id="B48">
  <title><p>Ketamine: A Neglected Therapy for Alzheimer Disease</p></title>
  <aug>
    <au><snm>Smalheiser</snm><fnm>N R.</fnm></au>
  </aug>
  <source>Frontiers in Aging Neuroscience</source>
  <pubdate>2019</pubdate>
  <volume>11</volume>
  <url>https://www.frontiersin.org/article/10.3389/fnagi.2019.00186</url>
</bibl>

<bibl id="B49">
  <title><p>Evaluation of the Neuroprotective Potential of N-Acetylcysteine for
  Prevention and Treatment of Cognitive Aging and Dementia.</p></title>
  <aug>
    <au><snm>Hara</snm><fnm>Y</fnm></au>
    <au><snm>McKeehan</snm><fnm>N</fnm></au>
    <au><snm>Dacks</snm><fnm>P A</fnm></au>
    <au><snm>Fillit</snm><fnm>H M</fnm></au>
  </aug>
  <source>J Prev Alzheimers Dis.</source>
  <pubdate>2017</pubdate>
</bibl>

<bibl id="B50">
  <title><p>SemMedDB: a PubMed-scale repository of biomedical semantic
  predications.</p></title>
  <aug>
    <au><snm>Kilicoglu</snm><fnm>H</fnm></au>
    <au><snm>Shin</snm><fnm>D</fnm></au>
    <au><snm>Fiszman</snm><fnm>M</fnm></au>
    <au><snm>Rosemblat</snm><fnm>G</fnm></au>
    <au><snm>Rindflesch</snm><fnm>TC.</fnm></au>
  </aug>
  <source>Bioinformatics</source>
  <pubdate>2012</pubdate>
  <volume>28(23)</volume>
</bibl>

<bibl id="B51">
  <title><p>Broad-coverage biomedical relation extraction with
  SemRep.</p></title>
  <aug>
    <au><snm>Kilicoglu</snm><fnm>H</fnm></au>
    <au><snm>Rosemblat</snm><fnm>G</fnm></au>
    <au><snm>M</snm><fnm>F</fnm></au>
    <au><snm>Shin</snm><fnm>D</fnm></au>
  </aug>
  <source>BMC Bioinformatics</source>
  <pubdate>2020</pubdate>
  <volume>21:1-28.</volume>
</bibl>

<bibl id="B52">
  <title><p>Extending the log-likelihood measure to improve collocation
  identification</p></title>
  <aug>
    <au><snm>McInnes</snm><fnm>B T</fnm></au>
  </aug>
  <source>Master's thesis</source>
  <publisher>Univerity of Minnesota, Minneapolis</publisher>
  <pubdate>2004</pubdate>
</bibl>

<bibl id="B53">
  <title><p>Learning Entity and Relation Embeddings for Knowledge Graph
  Completion</p></title>
  <aug>
    <au><snm>Lin</snm><fnm>Y</fnm></au>
    <au><snm>Liu</snm><fnm>Z</fnm></au>
    <au><snm>Sun</snm><fnm>M</fnm></au>
    <au><snm>Liu</snm><fnm>Y</fnm></au>
    <au><snm>Zhu</snm><fnm>X</fnm></au>
  </aug>
  <source>Proceedings of the Twenty-Ninth AAAI Conference on Artificial
  Intelligence</source>
  <publisher>Austin, Texas: AAAI Press</publisher>
  <series><title><p>AAAI'15</p></title></series>
  <pubdate>2015</pubdate>
  <fpage>2181–2187</fpage>
</bibl>

<bibl id="B54">
  <title><p>DGL-KE: Training Knowledge Graph Embeddings at Scale</p></title>
  <aug>
    <au><snm>Zheng</snm><fnm>D</fnm></au>
    <au><snm>Song</snm><fnm>X</fnm></au>
    <au><snm>Ma</snm><fnm>C</fnm></au>
    <au><snm>Tan</snm><fnm>Z</fnm></au>
    <au><snm>Ye</snm><fnm>Z</fnm></au>
    <au><snm>Dong</snm><fnm>J</fnm></au>
    <au><snm>Xiong</snm><fnm>H</fnm></au>
    <au><snm>Zhang</snm><fnm>Z</fnm></au>
    <au><snm>Karypis</snm><fnm>G</fnm></au>
  </aug>
  <source>Proceedings of the 43rd International ACM SIGIR Conference on
  Research and Development in Information Retrieval</source>
  <publisher>New York, NY, USA: Association for Computing Machinery</publisher>
  <series><title><p>SIGIR '20</p></title></series>
  <pubdate>2020</pubdate>
  <fpage>739–748</fpage>
</bibl>

<bibl id="B55">
  <title><p>Translating Embeddings for Modeling Multi-relational
  Data</p></title>
  <aug>
    <au><snm>Bordes</snm><fnm>A</fnm></au>
    <au><snm>Usunier</snm><fnm>N</fnm></au>
    <au><snm>Garcia Duran</snm><fnm>A</fnm></au>
    <au><snm>Weston</snm><fnm>J</fnm></au>
    <au><snm>Yakhnenko</snm><fnm>O</fnm></au>
  </aug>
  <publisher>Red Hook, NY, USA: Advances in Neural Information Processing
  Systems</publisher>
  <editor>C. J. C. Burges and L. Bottou and M. Welling and Z. Ghahramani and K.
  Q. Weinberger</editor>
  <pubdate>2013</pubdate>
  <volume>26</volume>
</bibl>

<bibl id="B56">
  <title><p>Embedding Entities and Relations for Learning and Inference in
  Knowledge Bases</p></title>
  <aug>
    <au><snm>Yang</snm><fnm>B</fnm></au>
    <au><snm>Yih</snm><fnm>W</fnm></au>
    <au><snm>X</snm><fnm>H</fnm></au>
    <au><snm>Gao</snm><fnm>J</fnm></au>
    <au><snm>Deng</snm><fnm>L</fnm></au>
  </aug>
  <pubdate>2015</pubdate>
</bibl>

<bibl id="B57">
  <title><p>Complex Embeddings for Simple Link Prediction</p></title>
  <aug>
    <au><snm>Trouillon</snm><fnm>T</fnm></au>
    <au><snm>Welbl</snm><fnm>J</fnm></au>
    <au><snm>Riedel</snm><fnm>S</fnm></au>
    <au><snm>Gaussier</snm><fnm>E</fnm></au>
    <au><snm>Bouchard</snm><fnm>G</fnm></au>
  </aug>
  <pubdate>2016</pubdate>
</bibl>

<bibl id="B58">
  <title><p>Literature Based Discovery: Models, methods, and trends</p></title>
  <aug>
    <au><snm>Henry</snm><fnm>S</fnm></au>
    <au><snm>McInnes</snm><fnm>B T</fnm></au>
  </aug>
  <source>Journal of Biomedical Informatics</source>
  <pubdate>2017</pubdate>
  <volume>74</volume>
  <fpage>20</fpage>
  <lpage>32</lpage>
  <url>https://www.sciencedirect.com/science/article/pii/S1532046417301909</url>
</bibl>

</refgrp>
} % end of \BMCxmlcomment
% for author-year bibliography (bmc-mathphys or spbasic)
% a) write to bib file (bmc-mathphys only)
% @settings{label, options="nameyear"}

% or include bibliography directly:
% \begin{thebibliography}
% \bibitem{b1}
% \end{thebibliography}

%%%%%%%%%%%%%%%%%%%%%%%%%%%%%%%%%%%
%%                               %%
%% Figures                       %%
%%                               %%
%% NB: this is for captions and  %%
%% Titles. All graphics must be  %%
%% submitted separately and NOT  %%
%% included in the Tex document  %%
%%                               %%
%%%%%%%%%%%%%%%%%%%%%%%%%%%%%%%%%%%

%%
%% Do not use \listoffigures as most will included as separate files

% \section*{Figures}
%   \begin{figure}[h!]
%   \caption{\csentence{Sample figure title.}
%       A short description of the figure content
%       should go here.}
%       \end{figure}

% \begin{figure}[h!]
%   \caption{\csentence{Sample figure title.}
%       Figure legend text.}
%       \end{figure}

%%%%%%%%%%%%%%%%%%%%%%%%%%%%%%%%%%%
%%                               %%
%% Tables                        %%
%%                               %%
%%%%%%%%%%%%%%%%%%%%%%%%%%%%%%%%%%%

%% Use of \listoftables is discouraged.
%%

%%%%%%%%%%%%%%%%%%%%%%%%%%%%%%%%%%%
%%                               %%
%% Additional Files              %%
%%                               %%
%%%%%%%%%%%%%%%%%%%%%%%%%%%%%%%%%%%

\section*{Additional Files}
\section*{Tables}
\begin{table}[h!]
\caption{Special Subjects and Objects Been Kept}
      \begin{tabular}{cc}\hline 
    Object/Subject Name  &   CUI \\
    \hline
    Alzheimer Disease, Early Onset	 & C0750901\\
    Alzheimer Disease, Late Onset	 & C0494463\\
    Focal Alzheimer's disease &	C0338450\\
    Familial Alzheimer's disease &	C0276496\\
    Alzheimer's disease treatment &	C1979617\\
    Alzheimer's disease antigen	 & C0051532\\
    Alzheimer's Disease	& C0002395\\
    Alzheimer Disease 11 &	C1853360\\
    Alzheimer Disease 14 &	C1970144\\
    Alzheimer Disease 16 &	C2677888\\
    Alzheimer Disease 7 &	C1853555\\
    Alzheimer Disease 8	 &C1846735\\
     Alzheimer disease, familial, type 3	& C1843013\\
     Alzheimer Vaccines	& C0949574\\\hline
      \end{tabular}
\end{table}

\end{backmatter}
\end{document}